\documentclass{article}

\PassOptionsToPackage{numbers, compress}{natbib}

\usepackage[final]{styles/neurips_2024}

\usepackage{amsmath}
\usepackage{multirow}
\usepackage{graphicx}
\usepackage{xcolor}
\usepackage{booktabs} 
\usepackage[colorlinks=true, allcolors=blue]{hyperref}
\usepackage{subfigure}

\usepackage{listings}
\title{MolMole: Molecule Mining from Scientific Literature}

\begin{document}

\begin{figure}[h]
\centering
\includegraphics[width=0.25\textwidth]{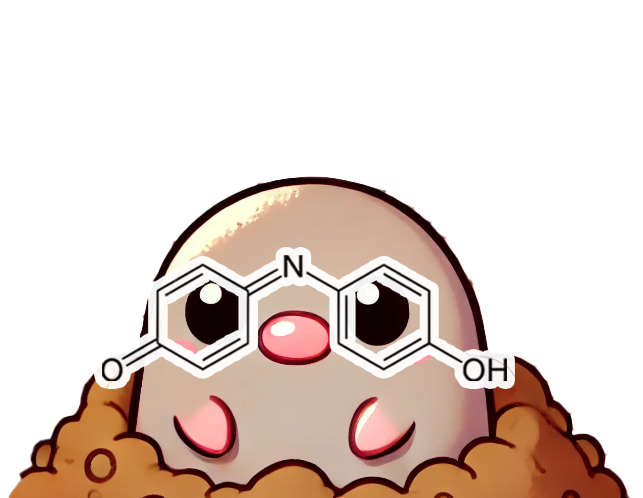}
\label{fig:mole}
\end{figure}
\vspace{-0.55cm}
\maketitle

\vspace{-2cm}

\begin{center}
\textbf{LG AI Research} \\
\href{https://lgai-ddu.github.io/molmole/}{\texttt{https://lgai-ddu.github.io/molmole/}}

\end{center}

\begin{abstract}

The extraction of molecular structures and reaction data from scientific documents is challenging due to their varied, unstructured chemical formats and complex document layouts. To address this, we introduce MolMole, a vision-based deep learning framework that unifies molecule detection, reaction diagram parsing, and optical chemical structure recognition (OCSR) into a single pipeline for automating the extraction of chemical data directly from page-level documents. Recognizing the lack of a standard page-level benchmark and evaluation metric, we also present a testset of 550 pages annotated with molecule bounding boxes, reaction labels, and MOLfiles, along with a novel evaluation metric. Experimental results demonstrate that MolMole outperforms existing toolkits on both our benchmark and public datasets. The benchmark testset will be publicly available, and the MolMole toolkit will be accessible soon through an interactive demo on the LG AI Research website. For commercial inquiries, please contact us at \href{mailto:contact_ddu@lgresearch.ai}{contact\_ddu@lgresearch.ai}.

\end{abstract}

\section{Introduction}

The rapid growth of scientific publications in chemistry and materials science has led to an overwhelming accumulation of molecular structure and reaction data. However, much of this valuable information remains embedded in unstructured formats, such as images, figures, and complex diagrams. Converting this data into machine-readable formats is essential for integrating it into public databases, enabling large-scale analysis, and accelerating research. Traditionally, this extraction process has been manual and time-consuming, requiring significant human effort and resources.

In recent years, several AI-driven frameworks have been developed for document-level molecular data extraction, with DECIMER \cite{rajan2023decimer} and OpenChemIE \cite{fan2024openchemie} being among the most prominent. DECIMER~\cite{rajan2023decimer} is the first publicly available framework to incorporate molecule segmentation, classification, and Optical Chemical Structure Recognition (OCSR). However, it lacks the ability to process reaction diagrams, limiting comprehensive chemical data extraction. In contrast, OpenChemIE~\cite{fan2024openchemie} achieves strong OCSR and reaction diagram parsing performance by leveraging multiple AI models. However, it relies on an external layout parser model~\cite{shen2021layoutparser} to crop document elements, which can lead to detection failures in complex layouts. 

In this work, we introduce \textbf{MolMole}, a vision-based deep learning toolkit for page-level molecular information extraction. Unlike existing frameworks, MolMole directly processes full document pages without requiring a layout parser, enabling efficient extraction of molecular structures and reaction data from complex scientific documents. It integrates molecule detection (ViDetect), reaction diagram parsing (ViReact) and OCSR (ViMore) into a unified workflow, allowing direct processing of page-level input.

Moreover, to enable systematic evaluation of document-level molecular information extraction, we curated a comprehensive page-level testset and introduced an evaluation metric tailored for this task. While datasets exist for tasks such as OCSR~\cite{rajan2020review}, molecule segmentation~\cite{rajan2021decimer}, and reaction diagram parsing~\cite{qian2023rxnscribe}, no unified benchmark assesses the full extraction pipeline, making direct performance comparison difficult. Our dataset, the first of its kind, consists of 550 annotated document pages, including 3,897 molecular structures and 1,022 reactions, providing a standardized framework for evaluating molecular data extraction from scientific literature.

\begin{figure}[t!]
\centering
\includegraphics[width=1.0\textwidth]{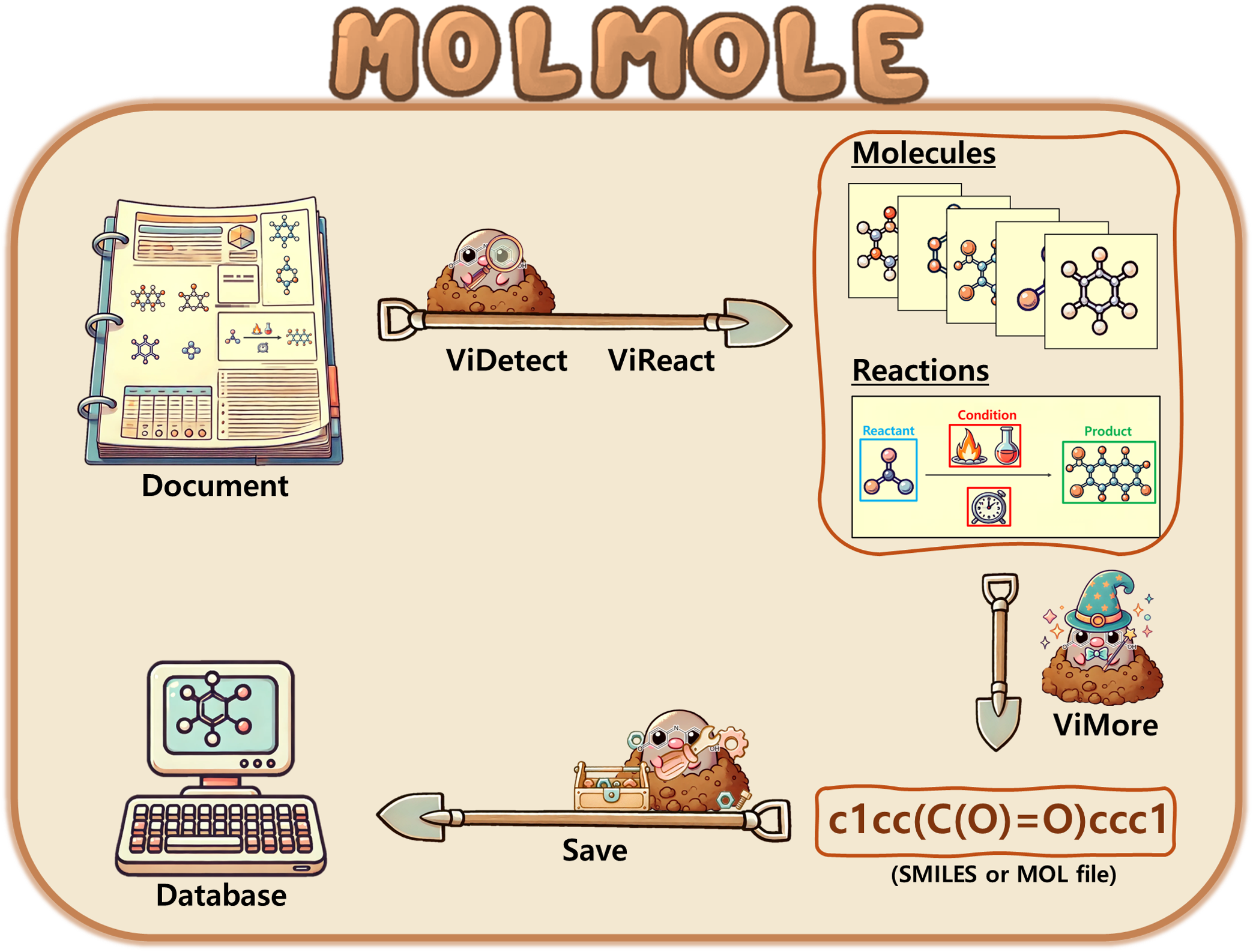}
\caption{MolMole pipeline. ViDetect detects molecular regions in document images, while ViReact extracts reactants, products, and conditions from reaction diagrams. ViMore processes the identified molecular structures, converting them into SMILES or MOLfiles.}
\label{fig:pipeline}
\end{figure}

In our experiments, MolMole outperformed existing toolkits, including OpenChemIE, DECIMER 2.0, and ReactionDataExtractor 2.0~\cite{wilary2023reactiondataextractor}, when evaluated on our new page-level benchmark dataset. It achieved F1 scores of 89.1\% and 86.8\% for the combined performance of molecular detection and OCSR, and 98.0\% and 97.0\% for page-level reaction diagram parsing on patents and articles, respectively. Additionally, on OCSR public benchmark, MolMole outperformed on three out of four datasets on molecule conversion accuracy. 

The following list summarizes the key contributions of \textbf{MolMole}:

\begin{itemize}
    \item MolMole offers an end-to-end framework for extracting chemical information at the page level by seamlessly combining molecule detection, reaction diagram parsing, and optical chemical structure recognition (OCSR) into a single pipeline. 
    \item We constructed a benchmark dataset for page-level evaluation and proposed a novel metric tailored to assess chemical information extraction across entire documents. 
    \item MolMole outperforms existing tools in accuracy, achieving state-of-the-art results on both our page-level benchmark and public OCSR evaluation datasets. 
\end{itemize}

\newpage

\section{MolMole Pipeline}

\autoref{fig:pipeline} illustrates MolMole workflow, where PDF documents are converted into PNG images and processed by ViDetect and ViReact in parallel. ViDetect identifies molecular structures by detecting bounding boxes, while ViReact parses reaction diagrams and extracts key components such as reactants, conditions, and products. Once molecular regions are identified, ViMore converts molecular images into formats like MOLfiles~\cite{dalby1992description} or SMILES~\cite{weininger1988smiles}. The final data can then be saved in various formats, including JSON or Excel. The following sections detail AI models that power MolMole.

\subsection{ViDetect for Molecule Detection}
ViDetect (Vision Detection) is an object detection model designed to predict bounding boxes for molecular structures in document images. Its architecture is derived from DINO~\cite{zhang2022dino} and is trained end-to-end on our private dataset. To enhance detection accuracy, all predicted bounding boxes undergo post-processing to remove overlapping proposals based on confidence scores and size constraints. 

Existing molecule detection models take different approaches, but each has limitations for large-scale data processing. DECIMER’s segmentation-based method~\cite{rajan2021decimer} is computationally expensive, while OpenChemIE’s MolDet~\cite{fan2024openchemie} uses an autoregressive approach that slows inference as the number of molecules increases. To overcome these inefficiencies, ViDetect adopts a DETR-based architecture~\cite{carion2020end}, balancing speed and accuracy for large-scale molecular data extraction. This allows efficient processing of vast molecular datasets without the drawbacks of segmentation or autoregressive methods.

\subsection{ViReact for Reaction Diagram Parsing}
ViReact (Vision Reaction) is a deep learning model designed to extract structured reaction information directly from page-level document images. It identifies key reaction components, such as reactants, conditions, and products, while also predicting their bounding box coordinates and entity types. ViReact follows RxnScribe~\cite{qian2023rxnscribe} architecture, where the encoder abstracts the input image into hidden representations, and the decoder generates structured reaction sequences in an autoregressive manner. During inference, post-processing refines predictions by correcting duplicates and removing empty entities. 

Existing models like ReactionDataExtractor 2.0 and RxnScribe are trained on cropped reaction diagrams, requiring an additional step to first detect and extract these regions using models such as layout parsers~\cite{shen2021layoutparser}. This extra preprocessing can introduce errors and limit adaptability to complex document layouts. In contrast, ViReact operates directly on full-page inputs, removing the need for such preprocessing. To support this approach, we developed a custom page-level dataset with detailed annotations, incorporating reaction diagrams from both articles and patents across diverse formatting styles and structures.

\subsection{ViMore for Optical Chemical Structure Recognition}
ViMore (Vision Molecule Recognition) is an OCSR model that converts molecular images into machine-readable formats such as MOLfiles or SMILES. It detects atom regions, recognizes atomic symbols, and predicts bond types, assembling this information into structured molecular representations through postprocessing. Trained end-to-end on a proprietary dataset, ViMore achieves high accuracy in molecular structure recognition.

Unlike generative models such as MolScribe~\cite{qian2023molscribe} and DECIMER Image Transformer~\cite{rajan2023decimer}, which directly translate molecular images into SMILES sequences, ViMore adopts a detection-based approach. By explicitly predicting atom- and bond-level information, it avoids hallucination errors, improves interpretability, and enables layout-aware MOLfile generation. Moreover, ViMore is readily extensible beyond the constraints of SMILES, allowing it to recognize polymer structures with bracket notations and detect wavy bonds commonly found in patents (\autoref{fig:layout_preserving}). 

ViMore also assigns prediction confidence levels—low, medium, or high—to help users assess the reliability of its outputs. Screenshots of ViMore’s predictions with corresponding confidence scores are shown in \autoref{fig:workflow_results} and \autoref{fig:workflow_molexpand}.
\section{Performance}
\subsection{Benchmark}

A key challenge in developing and evaluating page-level extraction from chemical literature is the lack of end-to-end benchmark dataset. While OCSR benchmarks exist, they focus solely on image-to-molecule conversion without evaluating molecule detection, which is critical for page-level performance. To bridge this gap, we constructed a custom dataset that simulates real-world scenarios where an entire PDF serves as input, requiring the extraction of relevant chemical information. 

The dataset includes detailed, manually curated annotations for three core tasks: molecule detection, reaction parsing, and molecule conversion. The dataset comprises a total of 550 pages from scientific articles and patents, selected to capture diverse molecular structures, reaction diagrams, and layout variations. Each page has a full annotation of molecular bounding boxes, reaction diagram components (such as reactants, conditions, and products), and corresponding molecular representations in MOLfile format, enabling end-to-end evaluation of the whole pipeline. \autoref{table:testset_stats} shows the curated testset statistics: number of pages, total number of molecules and total number of reactions. 
\vspace{-3mm}
\begin{table}[h!]
\centering
\caption{Testset Statistics}

\begin{tabular}{lcccccc}  
\toprule
Dataset & \# Pages &  \# Molecules &  \# Reactions   \\ \midrule
Patents & 300 & 2,482 & 728  \\ 
Articles & 250 & 1,415 & 294  \\
\bottomrule
\end{tabular}

\label{table:testset_stats}
\end{table}
\vspace{-5mm}

\subsection{Evaluation}

We evaluated MolMole mainly against two state-of-the-art chemical information extraction frameworks, DECIMER 2.0 and OpenChemIE, both of which offer end-to-end processing from PDFs to extracted data. Specifically, ViDetect is compared with DECIMER Segmentation and OpenChemIE’s MolDetect, ViMore with DECIMER Image Transformer and OpenChemIE’s MolScribe, and ViReact with OpenChemIE’s RxnScribe and ReactionDataExtractor 2.0. The installation of all models used for comparison followed the procedures detailed in their original publications.

\subsubsection{Page-level Molecule Detection and Recognition}
\label{subsec:page_level_eval}

This section presents the page-level evaluation results, encompassing three distinct assessments: (1) molecule detection performance, (2) molecule conversion performance using ground truth (GT) bounding boxes, and (3) the combined performance of molecule detection and molecule conversion. The first two evaluations (1) and (2) are conducted independently to assess the effectiveness of molecule detection and conversion separately, without being influenced by each other’s outcomes. In contrast, the third evaluation (3) aims to measure the overall performance of the entire pipeline, from molecule detection to conversion. 

We evaluate molecule detection performance using standard object detection metrics: Average Precision (AP), Average Recall (AR), and F1 score. Following the COCO evaluation protocol~\cite{lin2014microsoft}, AP and AR are computed by averaging over multiple IoU (Intersection over Union) thresholds, ranging from 0.50 to 0.95 in 0.05 increments. \autoref{table:detect} summarizes the molecule detection performance of DECIMER Segmentation, MolDetect, and ViDetect on the Patents and Articles testsets. The results indicate that ViDetect consistently outperforms both baseline models across all metrics and datasets. On the Articles test set, ViDetect achieves an AP of 0.928, AR of 0.949, and F1 score of 0.938, surpassing the next best model (DECIMER Segmentation) by a notable margin. Similarly, on the Patents testset, it attains an AP of 0.914, AR of 0.938, and F1 score of 0.926, again outperforming the other models. These improvements underscore ViDetect’s robustness and effectiveness in handling complex and diverse document layouts, particularly in real-world patent and scholarly article formats.

Second, \autoref{table:mol} presents the molecule conversion performance of DECIMER Image Transformer, MolScribe, and ViMore. To evaluate molecule conversion performance in isolation, molecular regions are extracted from the pages of Patents and Articles using ground truth bounding boxes. The predicted MOLfile is then compared with the ground truth MOLfile using SMILES matching accuracy and Tanimoto similarity. 

\begin{table}[h!]
\centering
\caption{Molecule detection performance on Patents and Articles.}
\resizebox{0.8\textwidth}{!}{
\begin{tabular}{lcccccc}
\toprule
 & \multicolumn{3}{c}{Patents} & \multicolumn{3}{c}{Articles}\\ 
\cmidrule(lr){2-4} \cmidrule(lr){5-7} 
 Models & AP & AR & F1 & AP & AR & F1  \\ 
\midrule 
DECIMER Segmentation~\cite{rajan2021decimer} & 0.891 & 0.930 & 0.910 & 0.839 & 0.896 & 0.867  \\ 
MolDetect~\cite{fan2024openchemie} & 0.796 & 0.841 & 0.818 & 0.764 & 0.820 & 0.791  \\ 
ViDetect (Ours) & \textbf{0.914} & \textbf{0.938} & \textbf{0.926} & \textbf{0.928} & \textbf{0.949} & \textbf{0.938}  \\ 

\bottomrule
\end{tabular}
}
\vspace{-2mm}
\label{table:detect}
\end{table}
\begin{table}[h!]
\centering
\caption{Molecule conversion performance on Patents and Articles.}
\resizebox{0.8\textwidth}{!}{
\begin{tabular}{lcccc}
\toprule
 & \multicolumn{2}{c}{Patents} & \multicolumn{2}{c}{Articles}\\ 
\cmidrule(lr){2-3} \cmidrule(lr){4-5} 
 Models & SMILES & Tanimoto & SMILES & Tanimoto  \\ 
\midrule 
DECIMER Image Transformer~\cite{rajan2023decimer} & .753 & .914 & .681 & .892 \\ 
MolScribe~\cite{qian2023molscribe} & .709 & .913 & .729 & .951 \\ 
ViMore (Ours) & \textbf{.900} & \textbf{.957} & \textbf{.880} & \textbf{.931} \\ 

\bottomrule
\end{tabular}
}
\label{table:mol}
\end{table}



\begin{table}[h!]
\centering
\caption{Combined performance of molecule detection to conversion on Patents and Articles.}
\resizebox{1.0\textwidth}{!}{
\begin{tabular}{lcccccc}
\toprule
 & \multicolumn{3}{c}{Patents} & \multicolumn{3}{c}{Articles}\\ 
\cmidrule(lr){2-4} \cmidrule(lr){5-7} 
 Models & Precision & Recall & F1 & Precision & Recall & F1  \\ 
\midrule 
DECIMER Segmentation + Image Transformer~\cite{rajan2023decimer} & .738 & .737 & .738 & .673 & .673 & .673 \\ 
MolDetect + MolScribe~\cite{fan2024openchemie} & .693 & .682 & .688 & .701 & .710 & .706 \\ 
ViDetect + ViMore (Ours) & \textbf{.895} & \textbf{.887} & \textbf{.891} & \textbf{.867} & \textbf{.868} & \textbf{.868} \\ 

\bottomrule
\end{tabular}
}
\vspace{1mm}
\label{table:detect_mol}
\end{table}

ViMore achieves the highest performance on both the Patents and Articles benchmarks. Specifically, it attains a SMILES matching accuracy of 90\% on Patents and 88\% on Articles, significantly outperforming all other baselines.

Finally, \autoref{table:detect_mol} presents the overall performance of the full pipeline, from molecule detection to conversion. To assess the combined performance, we modify the conventional object detection metrics by incorporating SMILES string matching into precision and recall. 

Given the definitions of precision and recall,

\( TP \), \( FP \), and \( FN \) are determined as follows:

\vspace{-2mm}
\begin{equation}
TP = \sum_{i=1}^{N} \mathbf{1} \left( \max_{j} IoU(B_{gt}^{(i)}, B_{pred}^{(j)}) \geq \tau \text{ and } SMILES_{gt}^{(i)} = f_{I \to S}(B_{pred}^{(j)}) \right)
\end{equation}

Here, \( B_{gt}^{(i)} \) and \( B_{pred}^{(j)} \) are ground truth and predicted bounding boxes, respectively. \( SMILES_{gt}^{(i)} \) is the SMILES string associated with \( B_{gt}^{(i)} \), while \( f_{I \to S}(B_{pred}^{(j)}) \) denotes the predicted SMILES string derived from \( B_{pred}^{(j)} \) through the molecular conversion model. The IoU threshold \( \tau \) is set to 0.5. A False Positive (FP) occurs when a predicted bounding box does not correspond to any GT box or when its associated SMILES string differs from the GT SMILES string, computed as \( FP = | B_{pred} | - TP \). A False Negative (FN) arises when a GT object is not detected by any prediction or when its predicted SMILES string differs from the GT SMILES string, given by \( FN = | B_{gt} | - TP \). Here, \( | B_{pred} | \) and \( | B_{gt} | \) are total predicted and GT bounding boxes, respectively.

The results show that the combination of ViDetect and ViMore achieves the highest Precision, Recall, and F1 score on both the Patents and Articles test sets. Specifically, on the Patents benchmark, ViDetect + ViMore attains a precision of 0.895, recall of 0.887, and F1 score of 0.891, substantially outperforming the combinations of DECIMER Segmentation + Image Transformer and MolDetect + MolScribe. On the Articles benchmark, ViDetect + ViMore also leads with a precision of 0.867, recall of 0.868, and F1 score of 0.868. Overall, these results confirm the strong performance of our proposed method across both document types.

\begin{table}[h!]
\centering
\caption{Reaction parsing performance on Patents and Articles.}
\resizebox{1\textwidth}{!}{
\begin{tabular}{lcccccccccccc}
\toprule
 & \multicolumn{6}{c}{Patents} & \multicolumn{6}{c}{Articles} \\ 
\cmidrule(lr){2-7} \cmidrule(lr){8-13} 
 Models & \multicolumn{2}{c}{Precision} & \multicolumn{2}{c}{Recall} & \multicolumn{2}{c}{F1} & \multicolumn{2}{c}{Precision} & \multicolumn{2}{c}{Recall} & \multicolumn{2}{c}{F1} \\  
\cmidrule(lr){2-3} \cmidrule(lr){4-5} \cmidrule(lr){6-7} \cmidrule(lr){8-9} \cmidrule(lr){10-11} \cmidrule(lr){12-13}
 & Soft & Hard & Soft & Hard & Soft & Hard & Soft & Hard & Soft & Hard & Soft & Hard \\  
\midrule 
ReactionDataExtractor2.0(w/o LP) ~\cite{wilary2023reactiondataextractor}  & 0.406 & 0.155 & 0.282 & 0.107 & 0.332 & 0.127 & 0.526 & 0.160 & 0.313 & 0.095 & 0.392 & 0.119 \\ 

ReactionDataExtractor2.0(w/ LP) ~\cite{wilary2023reactiondataextractor} & 0.463 & 0.212 & 0.370 & 0.169 & 0.411 & 0.188 & 0.630 & 0.264 & 0.500 & 0.211 & 0.557 & 0.234 \\ 
RxnScribe(w/o LP)~\cite{qian2023rxnscribe} & 0.826 & 0.496 & 0.817 & 0.489 & 0.822 & 0.490 & 0.856 & 0.525 & 0.803 & 0.497 & 0.829 & 0.510 \\ 
RxnScribe(w/ LP)~\cite{qian2023rxnscribe} & 0.818 & 0.549 & 0.691 & 0.464 & 0.749 & 0.503 & 0.853 & 0.578 & 0.721 & 0.493 & 0.781 & 0.532 \\ 
ViReact (Ours) & \textbf{0.983} & \textbf{0.928} & \textbf{0.977} & \textbf{0.922} & \textbf{0.980} & \textbf{0.925} & \textbf{0.966} & \textbf{0.842} & \textbf{0.973} & \textbf{0.850} & \textbf{0.970} & \textbf{0.846} \\ 
\bottomrule
\end{tabular}
}
\vspace{1mm}
\label{table:react}
\end{table}

\begin{table}[h!]
\centering
\caption{OCSR performance on public benchmarks. InChI and SMILES refer to exact match accuracy based on InChI keys and SMILES strings, respectively.}
\resizebox{\textwidth}{!}{
\begin{tabular}{lcccccccc}
\toprule
 & \multicolumn{2}{c}{CLEF} & \multicolumn{2}{c}{JPO} & \multicolumn{2}{c}{UOB} & \multicolumn{2}{c}{USPTO}\\ 
\cmidrule(lr){2-3} \cmidrule(lr){4-5} \cmidrule(lr){6-7} \cmidrule(lr){8-9} 
 Models & InChI & SMILES  & InChI & SMILES  & InChI & SMILES  & InChI & SMILES  \\ 
\midrule 
DECIMER Image Transformer~\cite{rajan2023decimer} & .720 & .715  & .664 & .667  & \textbf{.987} & \textbf{.901}  & .630 & .608  \\
MolScribe~\cite{qian2023molscribe} & .796 & .830  & .753 & .756  & .983 & .896 & .934 & .935  \\
MolGrapher~\cite{morin2023molgrapher} & .496 & .493  & .556 & .560  & .950 & .869  & .639 & .635 \\

ViMore (ours) & \textbf{.853} & \textbf{.875}  & \textbf{.815} & \textbf{.815}  & .964 & .879 & \textbf{.938} & \textbf{.938}  \\

\bottomrule
\end{tabular}
}
\vspace{1mm}
\label{table:ocsr_public}
\end{table}

\subsubsection{Page-level Reaction Diagram Parsing}

To evaluate the performance of our reaction diagram parsing system, we adopt the hard match and soft match evaluation metrics proposed in RxnScribe. Predictions are compared against ground truth reactions using bounding box overlap, measured by Intersection over Union (IoU), where a match is considered successful if the highest IoU score exceeds 0.5. The soft match method evaluates only molecular entities, disregarding text labels and not differentiating between reactants and reagents, which helps account for visually ambiguous molecules near reaction arrows. In contrast, the hard match method requires the correct identification of all reaction components, including reactants, conditions, and products, with any misclassification resulting in an incorrect match. For both evaluation methods, we compute precision, recall, and F1 scores to quantify performance. For formal metric definitions and equations, we refer readers to RxnScribe.

\label{subsec:page_level_eval_reaction}

Since both RxnScribe and ReactionDataExtractor 2.0 are trained on isolated reaction diagrams, we apply a layout-aware preprocessing step using a layout parser to extract individual diagrams from full-page documents. In particular, RxnScribe is integrated into the OpenChemIE framework, which includes a layout parser module that crops diagram regions prior to prediction. To ensure a fair comparison, we adopt the same approach for ReactionDataExtractor 2.0. In our experiments, we report results for both versions of these models—w/ LP (with layout parser) and w/o LP (without layout parser)—to assess the impact of this preprocessing step. Our model, ViReact, by contrast, processes full page-level documents directly, without requiring external layout parsing.

\autoref{table:react} shows the performance of ReactionDataExtractor 2.0, RxnScribe and ViReact. ViReact outperforms all baseline models across all metrics and evaluation settings. On the Patents test set, ViReact achieves the highest F1 scores of 0.980 (soft) and 0.925 (hard), compared to the next best model, RxnScribe (w/o LP), which reaches 0.822 (soft) and 0.490 (hard). A similar trend is observed on the Articles test set, where ViReact attains F1 scores of 0.970 (soft) and 0.846 (hard), again surpassing the other models. Interestingly, RxnScribe (w/o LP) outperforms RxnScribe (w/ LP) in both Patents and Articles. This suggests that the layout parser module used in the OpenChemIE framework may introduce errors during diagram cropping, such as missed or incorrectly localizing regions, which negatively affect overall system performance. In contrast, ViReact’s direct page-level processing enables more reliable parsing, even in documents with complex layouts.

\subsubsection{OCSR Public Benchmark Evaluation}

This section evaluates molecule conversion models using publicly available OCSR benchmarks. We compare ViMore with state-of-the-art methods—DECIMER Image Transformer, MolScribe, and MolGrapher~\cite{morin2023molgrapher}—by conducting experiments on four standard benchmark datasets: USPTO, UOB, CLEF, and JPO~\cite{rajan2020review}, which contain 5719, 5740, 992, and 450 images, respectively.

The performance of each method is evaluated based on exact matching accuracy. Recognition correctness is determined by comparing the predicted molecules to the ground truth using both InChI keys~\cite{heller2015inchi} and SMILES representations. The results are summarized in \autoref{table:ocsr_public}. ViMore achieves the highest accuracy on three out of the four benchmarks—CLEF, JPO, and USPTO—recording InChI matching accuracies of 85.3\%, 81.5\%, and 93.8\%, respectively. Notably, it shows strong performance even on challenging datasets such as JPO.
\section{Discussion}
This section highlights the qualitative strengths of the MolMole framework that may not be fully captured through quantitative metrics.

\begin{figure}[t!]
\centering
\includegraphics[width=1.0\textwidth]{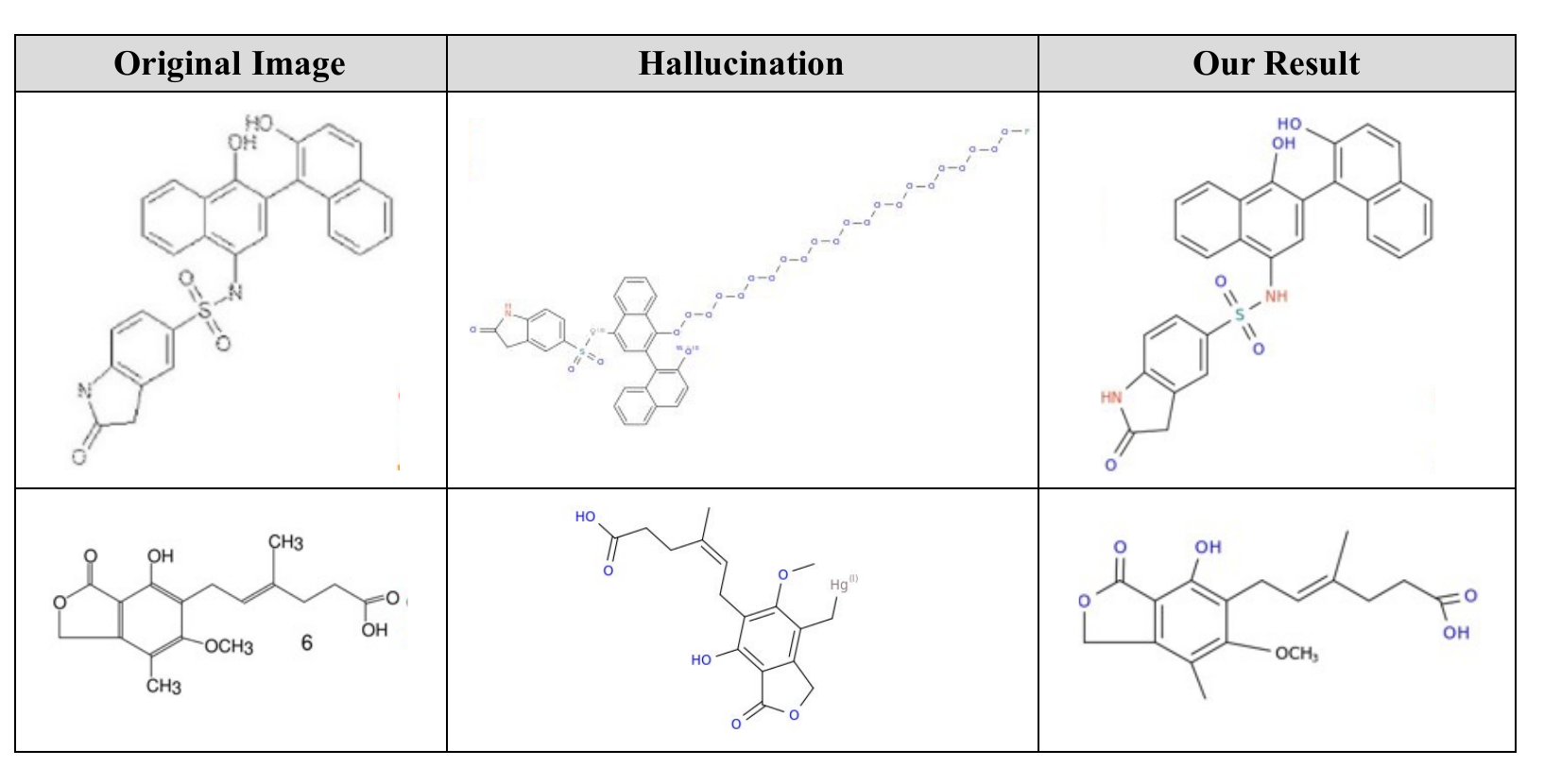}
\caption{Hallucination effects from generative models: repeated SMILES generation (top) and incorrect chemical bias (bottom).}
\label{fig:molscribe_ours}
\end{figure}

\paragraph{Reliable Recognition without Hallucination} 
Generative models such as MolScribe and vision-based models like ViMore differ fundamentally in their approach to molecular structure recognition. As illustrated in \autoref{fig:molscribe_ours}, generative models are prone to hallucination, producing unrealistic molecular structures or incorrect predictions due to biases toward specific chemical patterns. In contrast, ViMore explicitly detects atoms and bonds from the input, effectively mitigating hallucinations and structural biases. This leads to more interpretable and accurate extraction of molecular structures. Furthermore, SMILES-based generative models are limited to structures expressible within the SMILES syntax. In contrast, ViMore can handle structures beyond SMILES, such as polymers with wavy lines.

\begin{figure}[t!]
\centering
\includegraphics[width=1.0\textwidth]{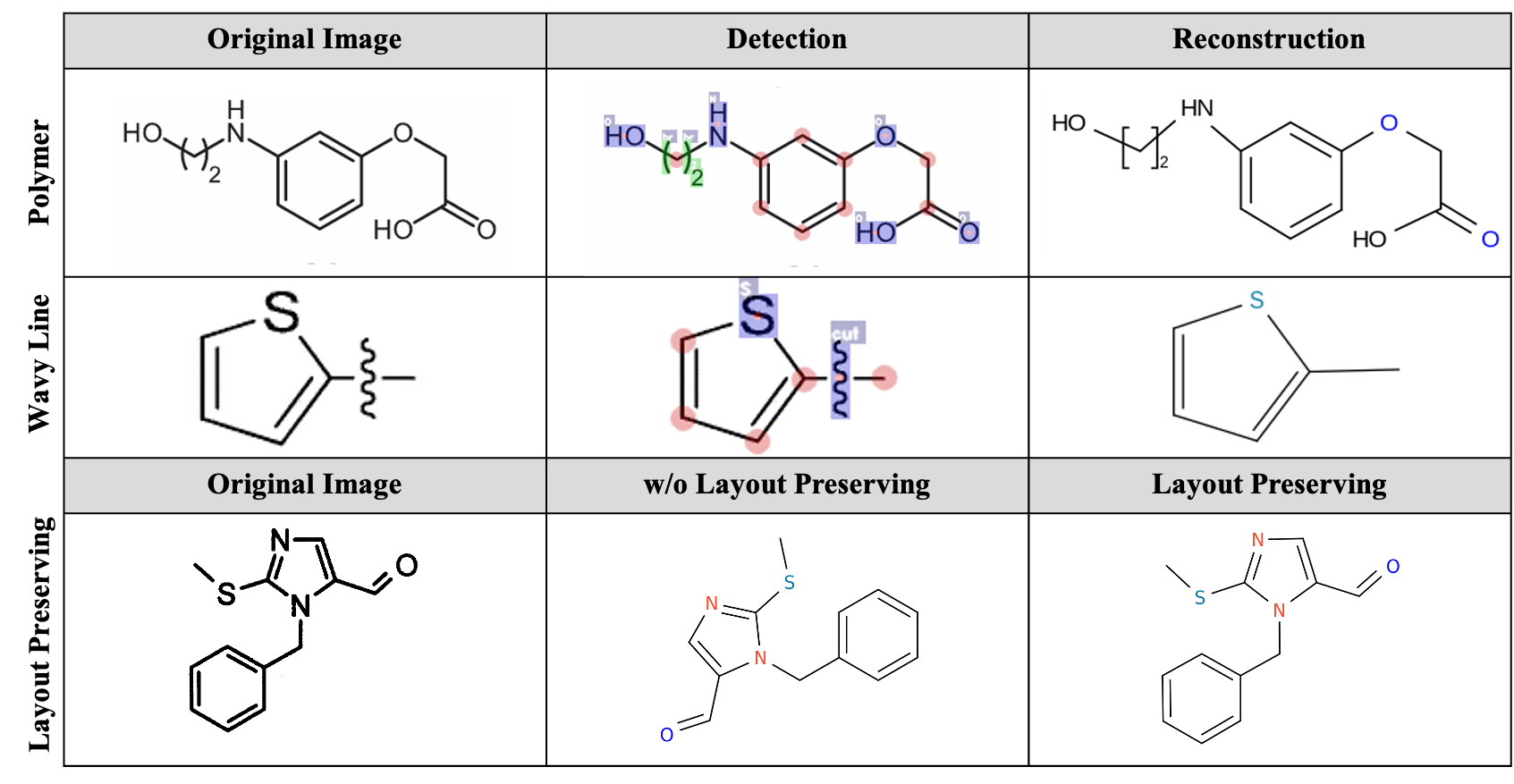}
\caption{ViMore results of Polymer (top) and Wavy line (middle). ViMore preserves the molecule coordinates from the image (bottom).}
\label{fig:layout_preserving}
\end{figure}

\paragraph{Layout-preserving MOL} 
ViMore generates layout-preserved MOLfiles that accurately retain the structure of the original image. This is a key advantage over existing OCSR models: for instance, the DECIMER Image Transformer does not include atomic position data, while MolScribe often fails to generate accurate coordinates. In contrast, ViMore leverages its detection-based architecture to produce accurate MOLfiles that closely mirror the original image. As illustrated in \autoref{fig:layout_preserving}, this not only simplifies the verification process but also enables quick and efficient edits when necessary.

\paragraph{Polymer and Wavy line }
\label{paragraph:polymer_wavy}
Polymers are typically depicted using brackets accompanied by a number, indicating the repetition of the enclosed substructure. Existing OCSR models often struggle to interpret this notation accurately. In contrast, ViMore reliably identifies both the brackets and the associated repetition count, enabling precise structural conversion. Additionally, in patent documents, a wavy line denotes a position where a variable substructure can be attached. Existing models often misinterpret wavy lines as single or other types of bonds. ViMore, however, correctly distinguishes wavy lines as separate graphical elements and generates MOLfiles with the wavy line appropriately excluded, resulting in a faithful molecular representation. These capabilities are illustrated in \autoref{fig:layout_preserving}.

\begin{figure}[t!]
\centering
\includegraphics[width=1\textwidth]{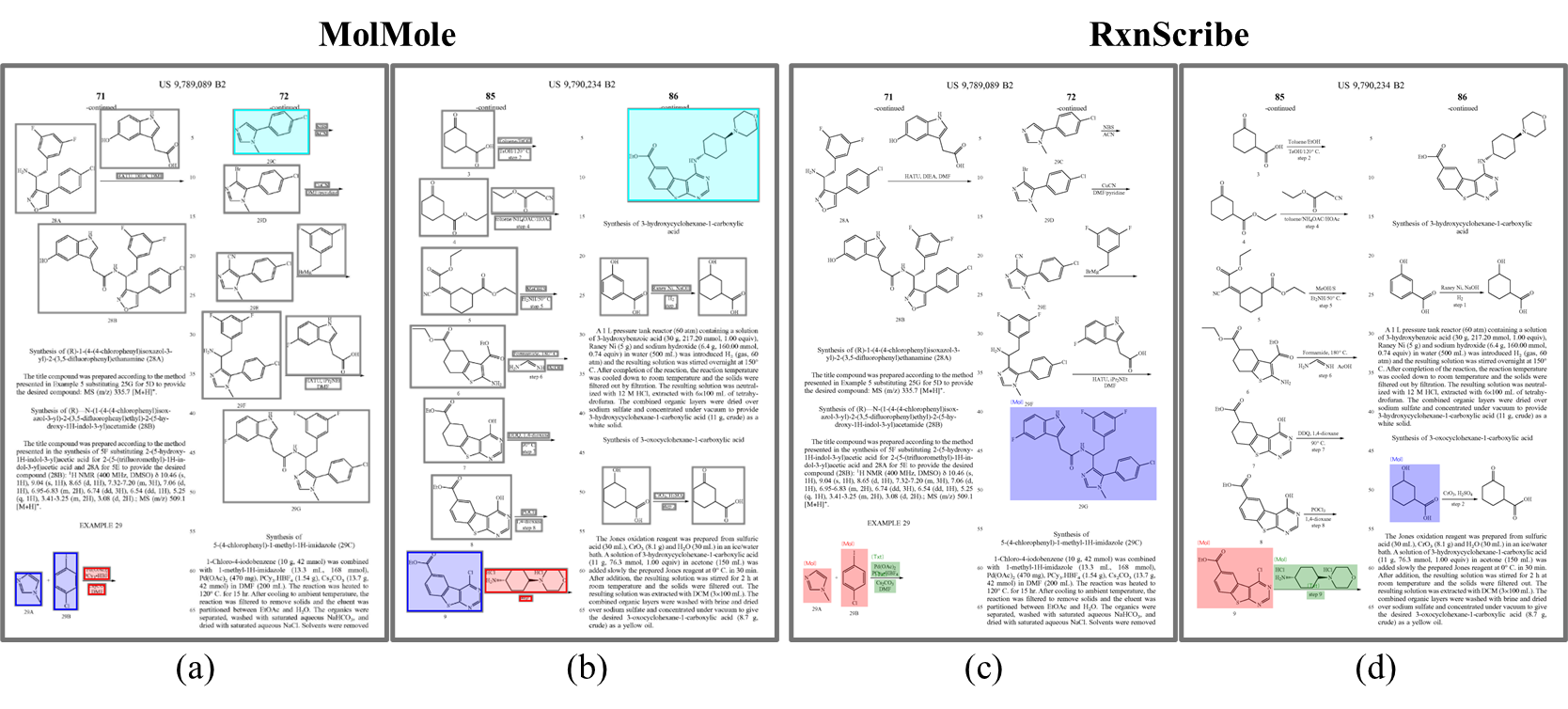}
\caption{Examples of reaction extraction in two-column documents. (a) and (b) show MolMole successfully extracting reaction information that spans from the first to the second column. In contrast, (c) and (d) show results from RxnScribe, which fails to capture the full reaction due to its reliance on cropped, isolated diagrams.}
\label{fig:2column}
\end{figure}

\paragraph{Two-Column Reaction Parsing}

A key distinction between MolMole and existing reaction parsing models, such as ReactionDataExtractor 2.0 and RxnScribe, lies in their ability to handle complex document layouts. As shown in \autoref{fig:2column}, MolMole accurately extracts reaction information even when it spans from the first to the second column in two-column documents—a scenario where other models typically struggle. This limitation arises because existing models are trained on isolated reaction diagrams rather than full document pages, and they rely on layout parsers to detect and crop diagrams before processing. In contrast, MolMole’s direct page-level processing offers a significant advantage, making it more reliable for extracting reaction information from complex scientific literature.

\section{Conclusion}
In this work, we introduce MolMole, a vision-based deep learning framework for extracting molecular structures and reaction data directly from scientific documents. Unlike existing approaches, MolMole processes entire document pages, integrating molecule detection, reaction parsing and OCSR into a unified pipeline for seamless end-to-end extraction. To support systematic evaluation, we present a new page-level benchmark dataset and a dedicated evaluation metric for document-level molecular data extraction. Experimental results demonstrate that MolMole outperforms existing toolkits on our benchmark dataset while achieving competitive performance across multiple OCSR benchmarks.

Beyond accuracy, MolMole introduces key advantages over existing models, including improved interpretability through its vision-based approach, layout-preserving MOLfiles, enhanced polymer and wavy line recognition, and robust reaction parsing in complex layouts such as two-column documents. Moving forward, we aim to further enhance MolMole’s ability to handle complex molecular representations and expand dataset coverage to improve generalizability. As the demand for automated molecular data extraction continues to grow, MolMole aims to drive AI-driven discoveries in chemistry and cheminformatics.

\section{Appendix}

\subsubsection{Contributors}
Core Contributors: 
Sehyun Chun, Jiye Kim, Ahra Jo, Yeonsik Jo, Seungyul Oh, Seungjun Lee, Kwangrok Ryoo, Jongmin Lee, Seung Hwan Kim, Byung Jun Kang, Soonyoung Lee, Jun Ha Park, Chanwoo Moon, Jiwon Ham, Haein Lee, Heejae Han, Jaeseung Byun, Soojong Do, Minju Ha, Dongyun Kim

Contributors: Kyunghoon Bae, Woohyung Lim, Edward Hwayoung Lee, Yongmin Park, Jeongsang Yu, Gerrard Jeongwon Jo, Yeonjung Hong, Kyungjae Yoo, Sehui Han, Jaewan Lee, Changyoung Park, Kijeong Jeon, Sihyuk Yi

\subsubsection{Qualitative Results}
\begin{figure}[h!]
  \centering
  \fbox{\includegraphics[width=0.48\linewidth]{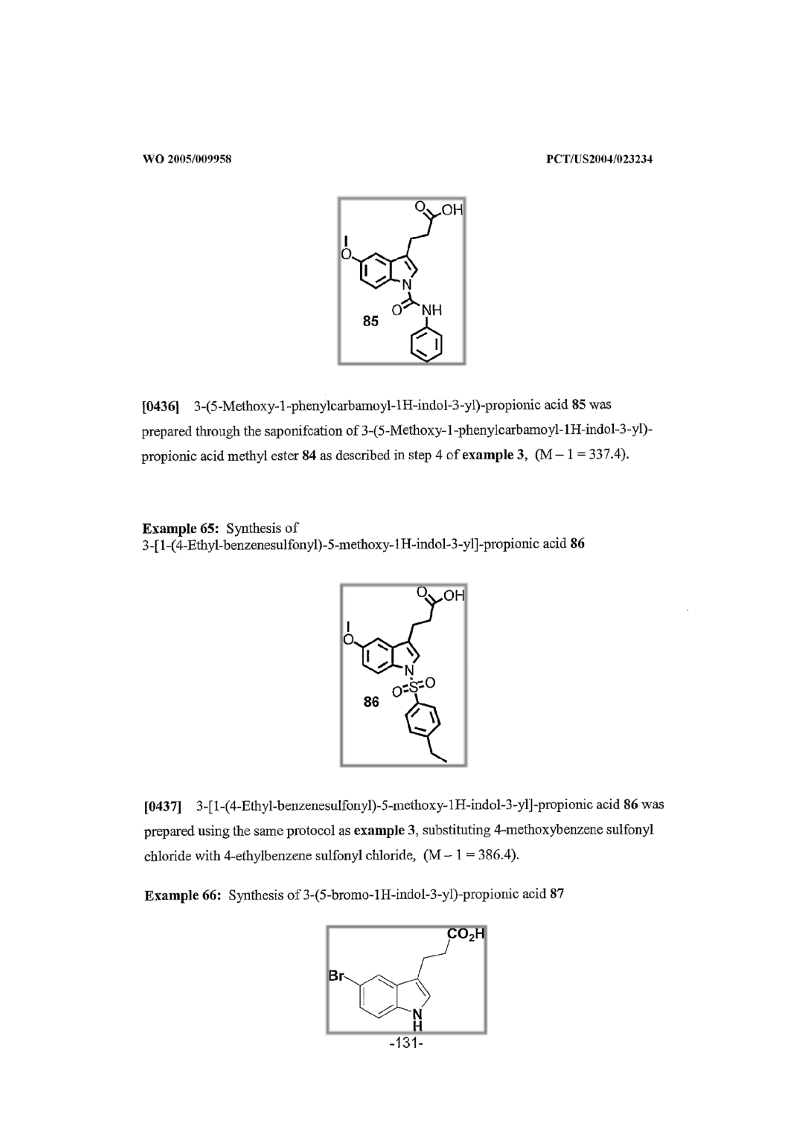}}
  \hspace{0.01\linewidth}
  \fbox{\includegraphics[width=0.48\linewidth]{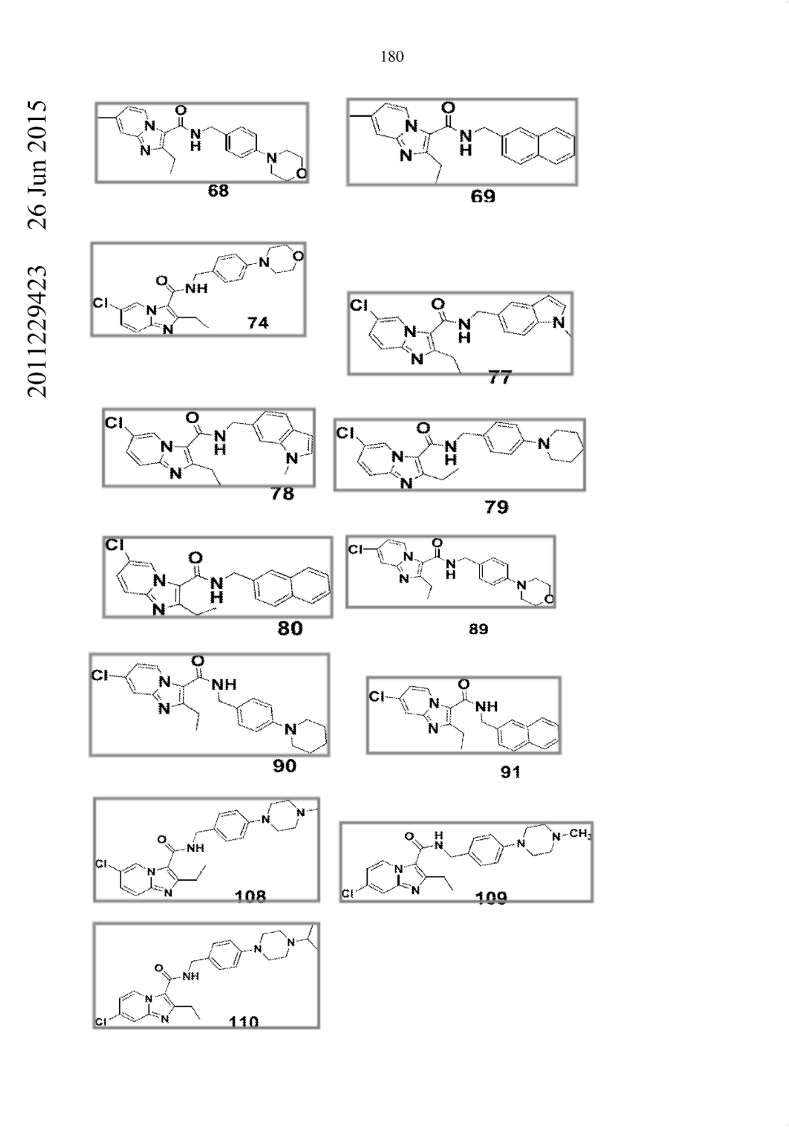}}\\[0.5ex]
  \fbox{\includegraphics[width=0.48\linewidth]{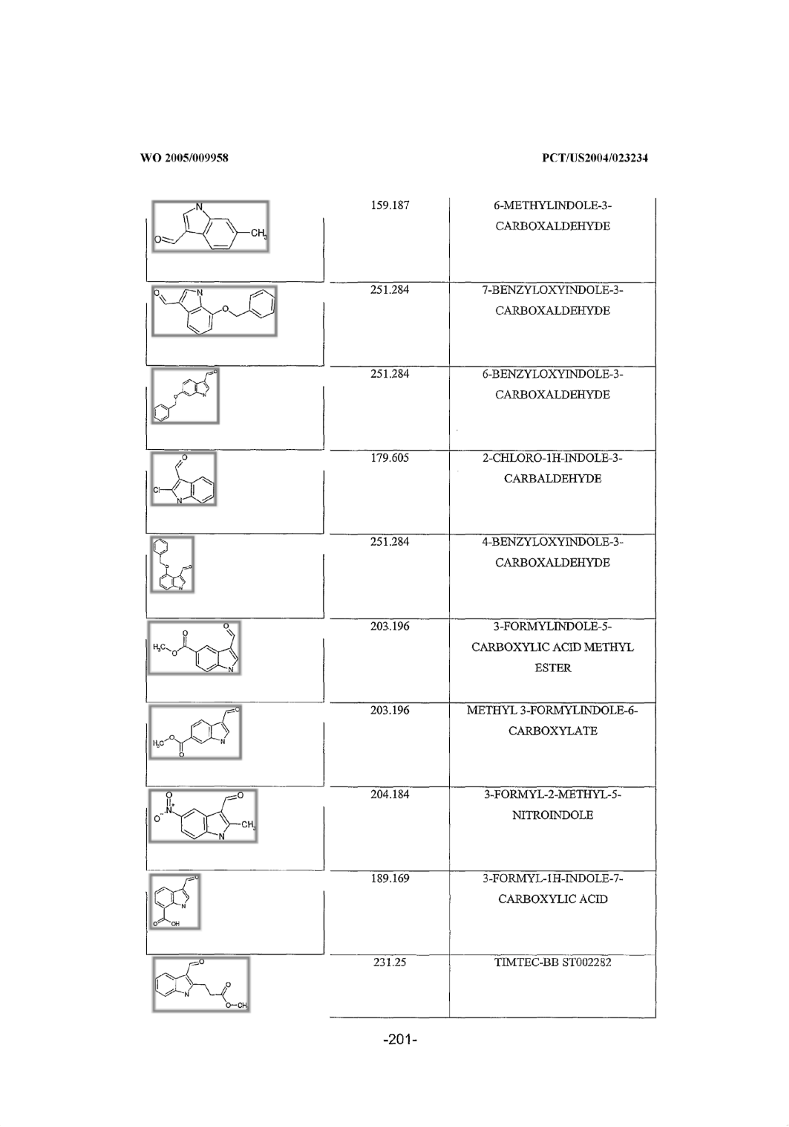}}
  \hspace{0.01\linewidth}
  \fbox{\includegraphics[width=0.48\linewidth]{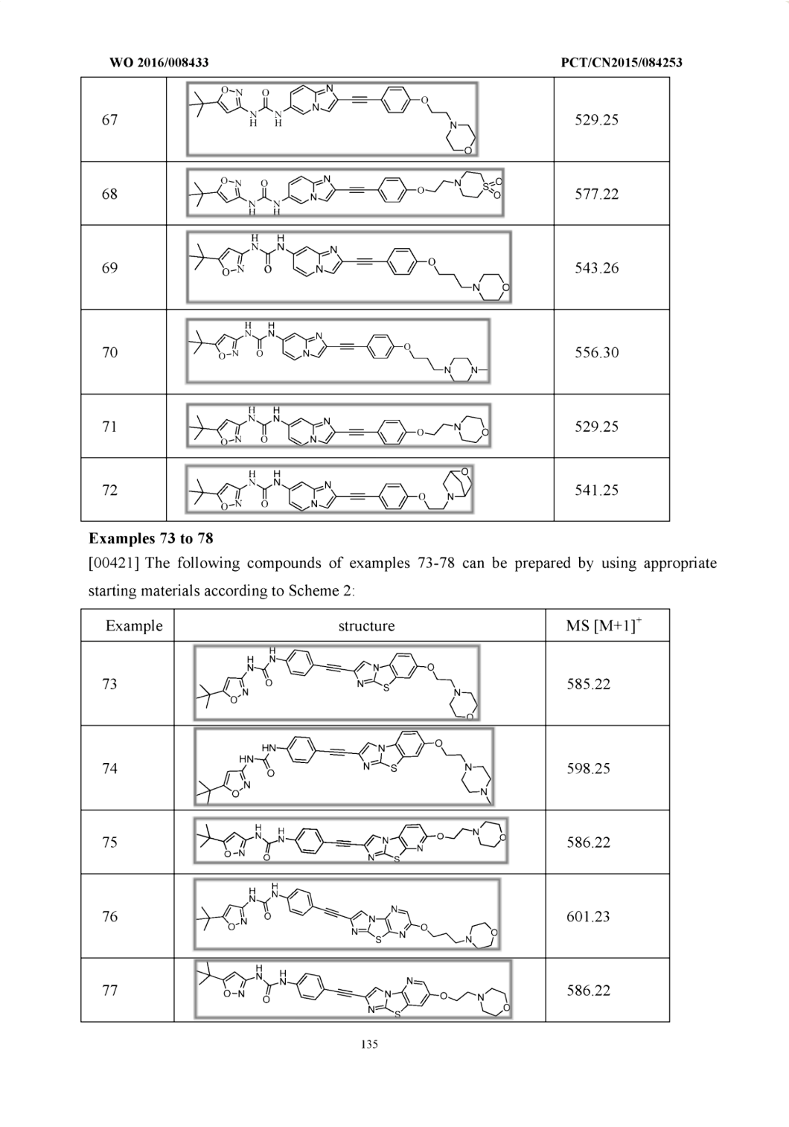}}
  \caption{Sample ViDetect results from our testset.}
  \label{fig:videtect_testset}
\end{figure}

\begin{figure}[h!]
  \centering
  \fbox{\includegraphics[width=0.48\linewidth]{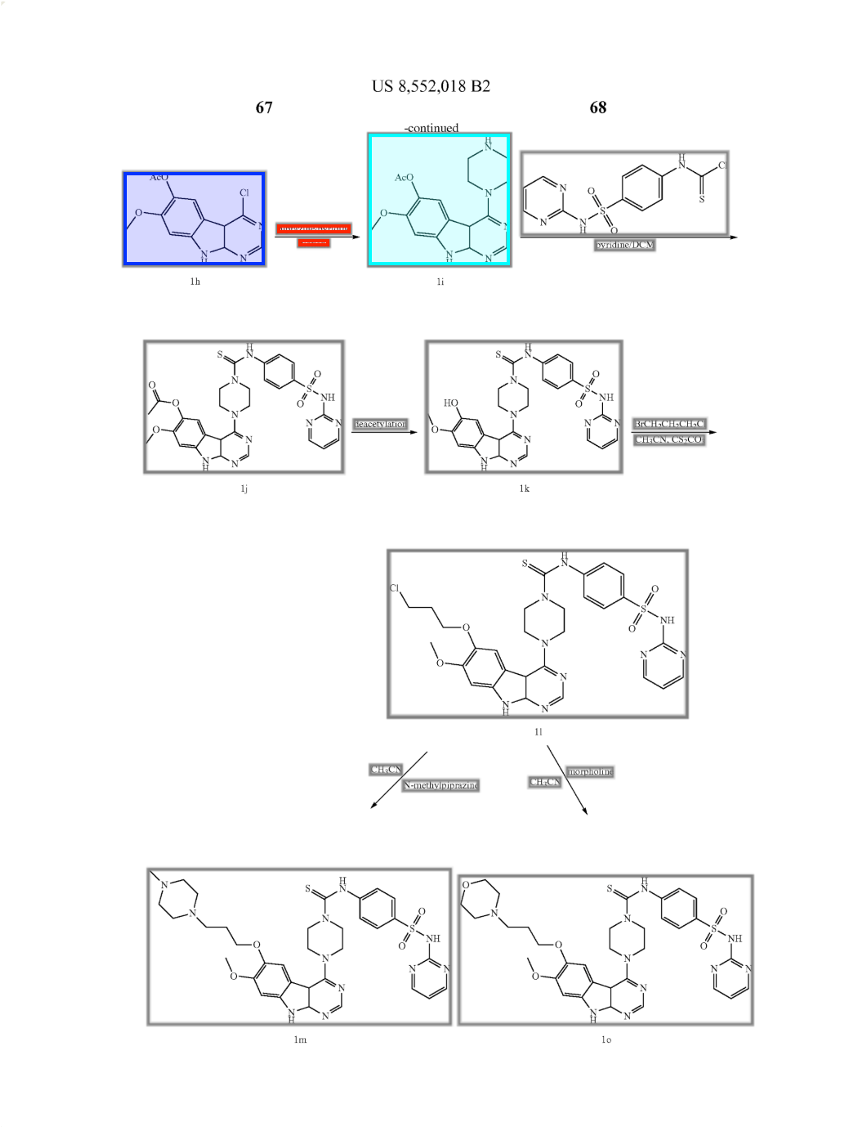}}
  \hspace{0.01\linewidth}
  \fbox{\includegraphics[width=0.48\linewidth]{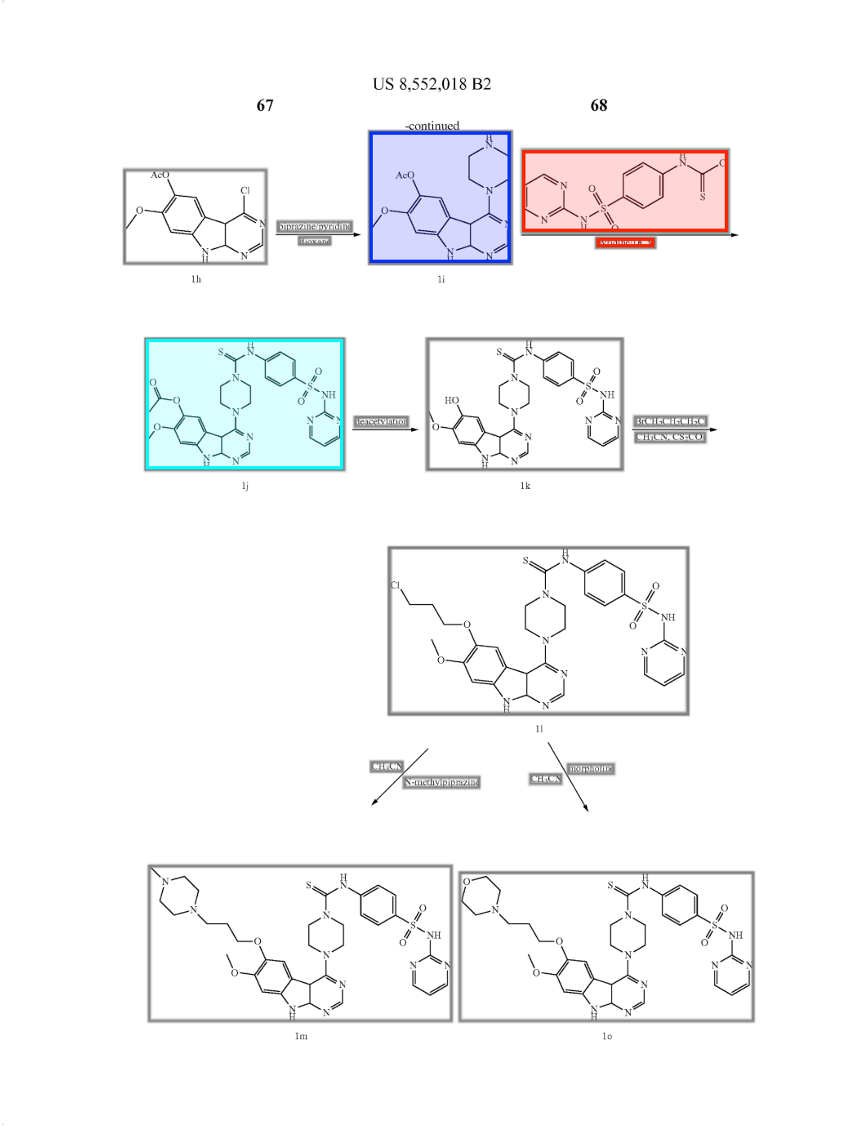}}\\[0.5ex]
  \fbox{\includegraphics[width=0.48\linewidth]{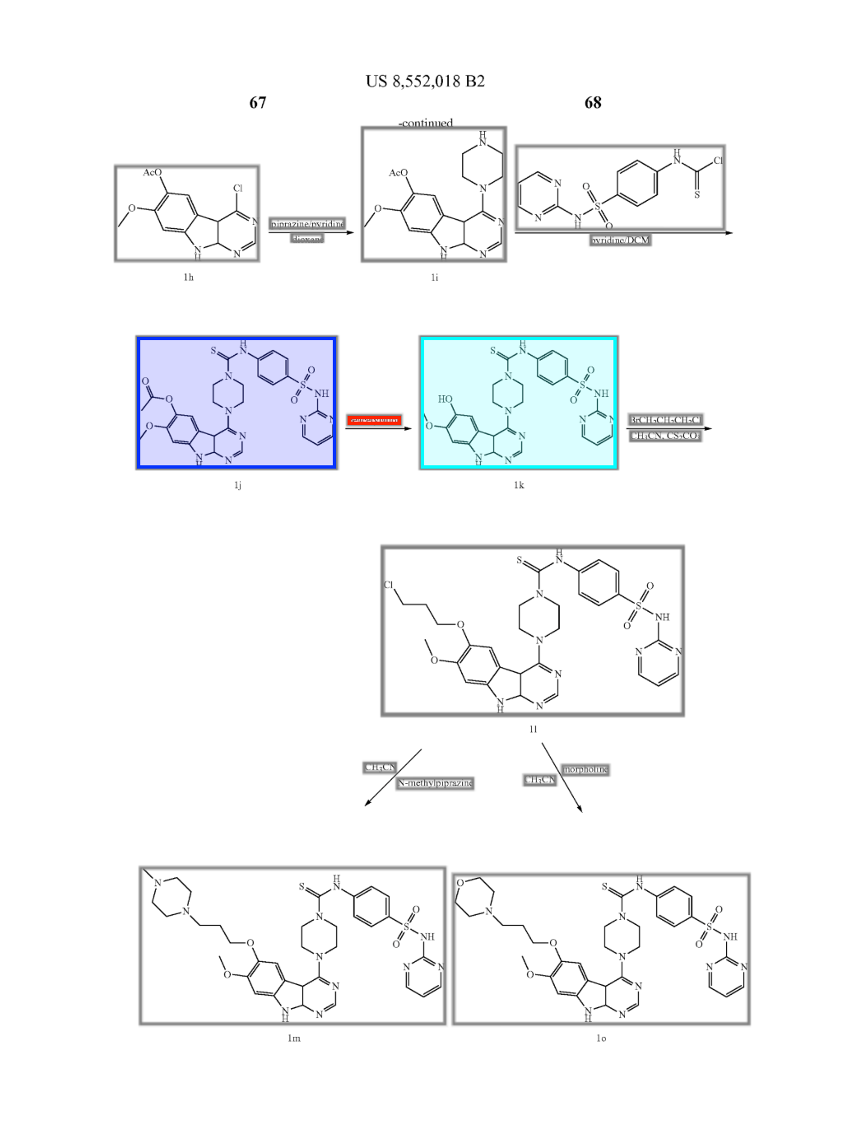}}
  \hspace{0.01\linewidth}
  \fbox{\includegraphics[width=0.48\linewidth]{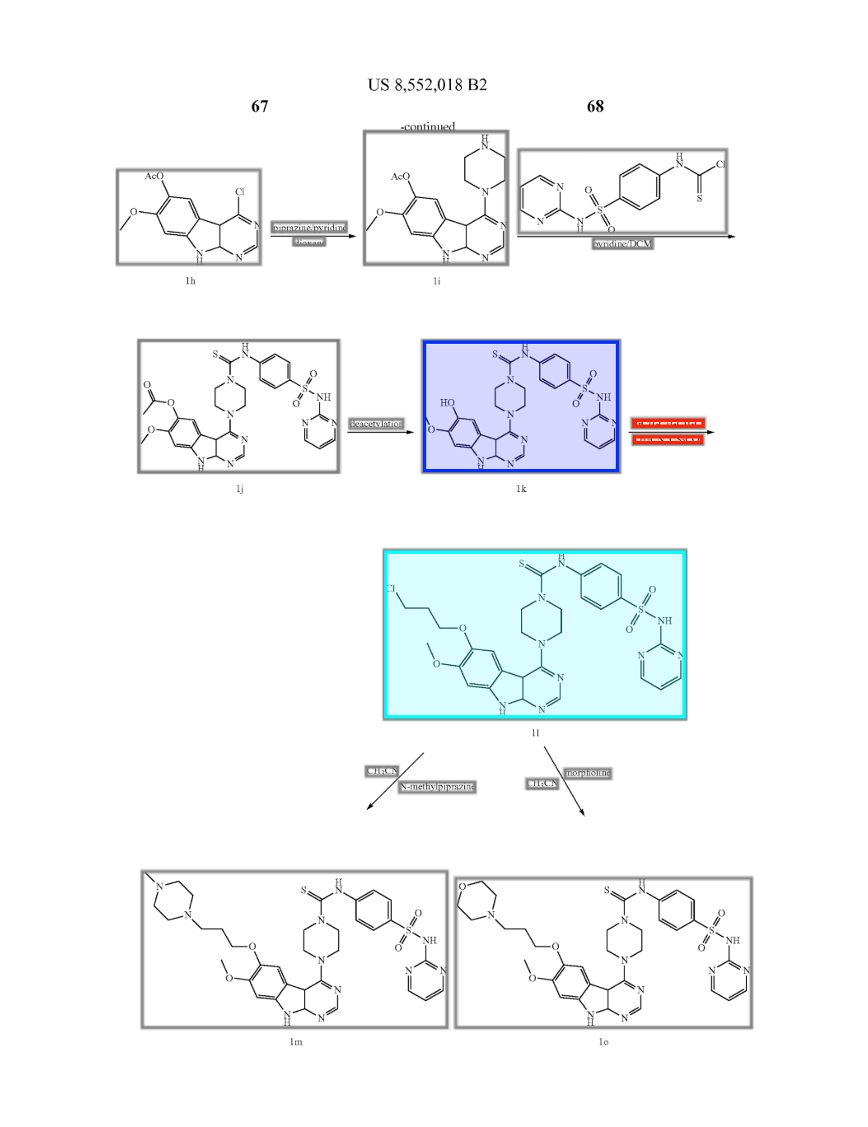}}
  \caption{Sample ViReact results from our testset. Dark blue, red and light blue indicate reactants, conditions and products, respectively.}
  \label{fig:vireact_testset}
\end{figure}

\begin{figure}[h!]
  \centering
  \fbox{\includegraphics[width=0.48\linewidth]{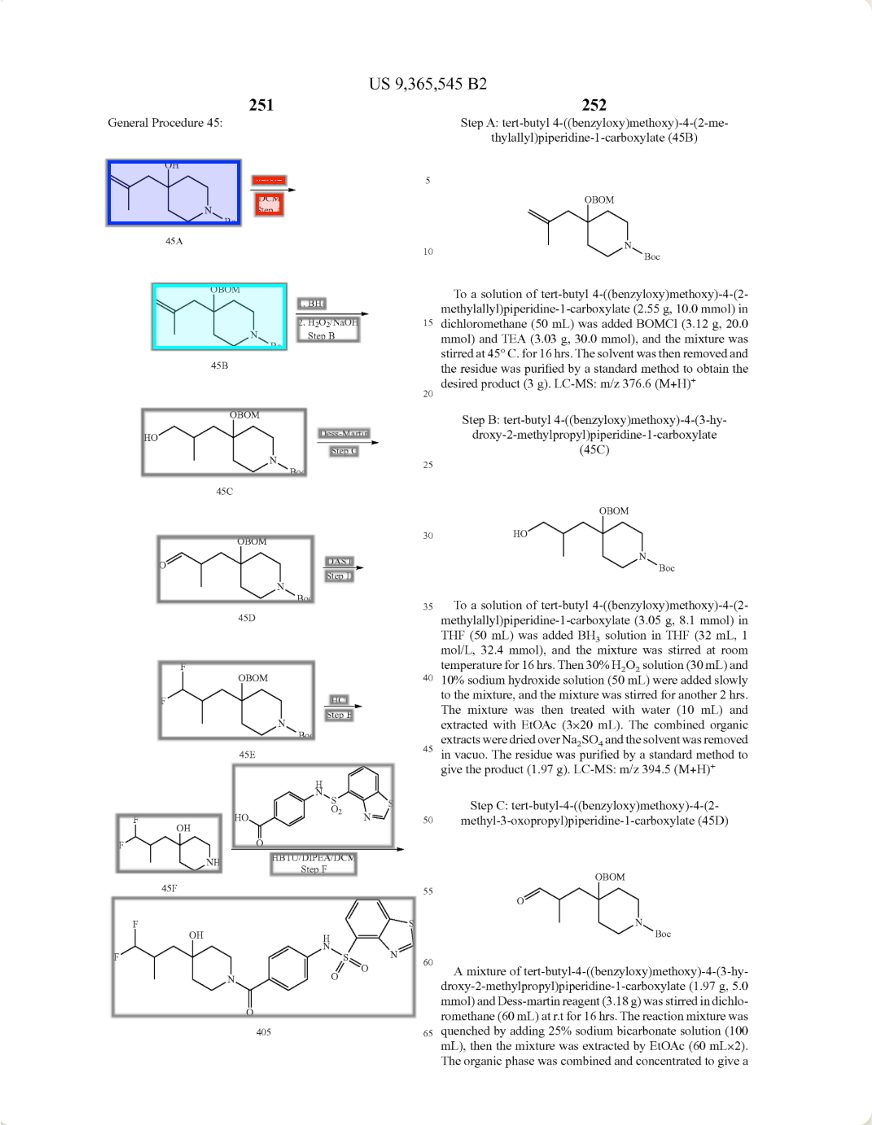}}
  \hspace{0.01\linewidth}
  \fbox{\includegraphics[width=0.48\linewidth]{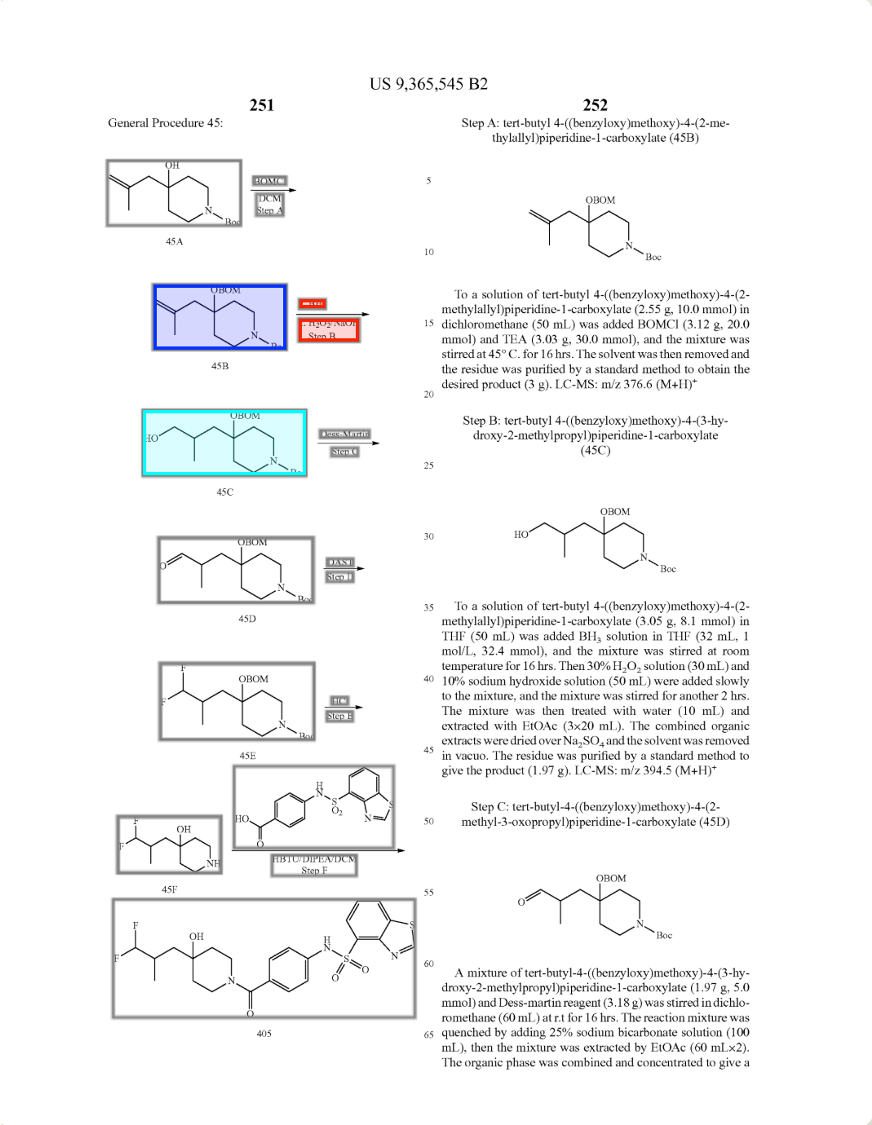}}\\[0.5ex]
  \fbox{\includegraphics[width=0.48\linewidth]{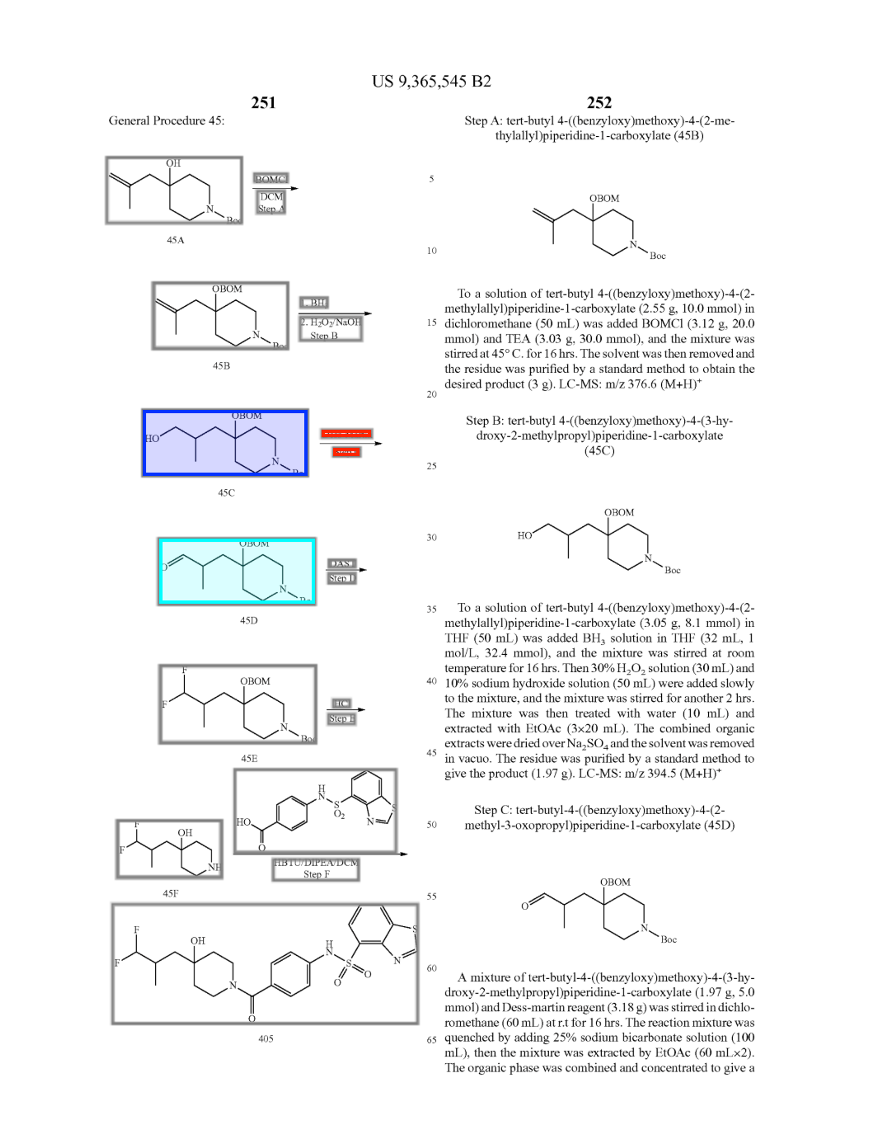}}
  \hspace{0.01\linewidth}
  \fbox{\includegraphics[width=0.48\linewidth]{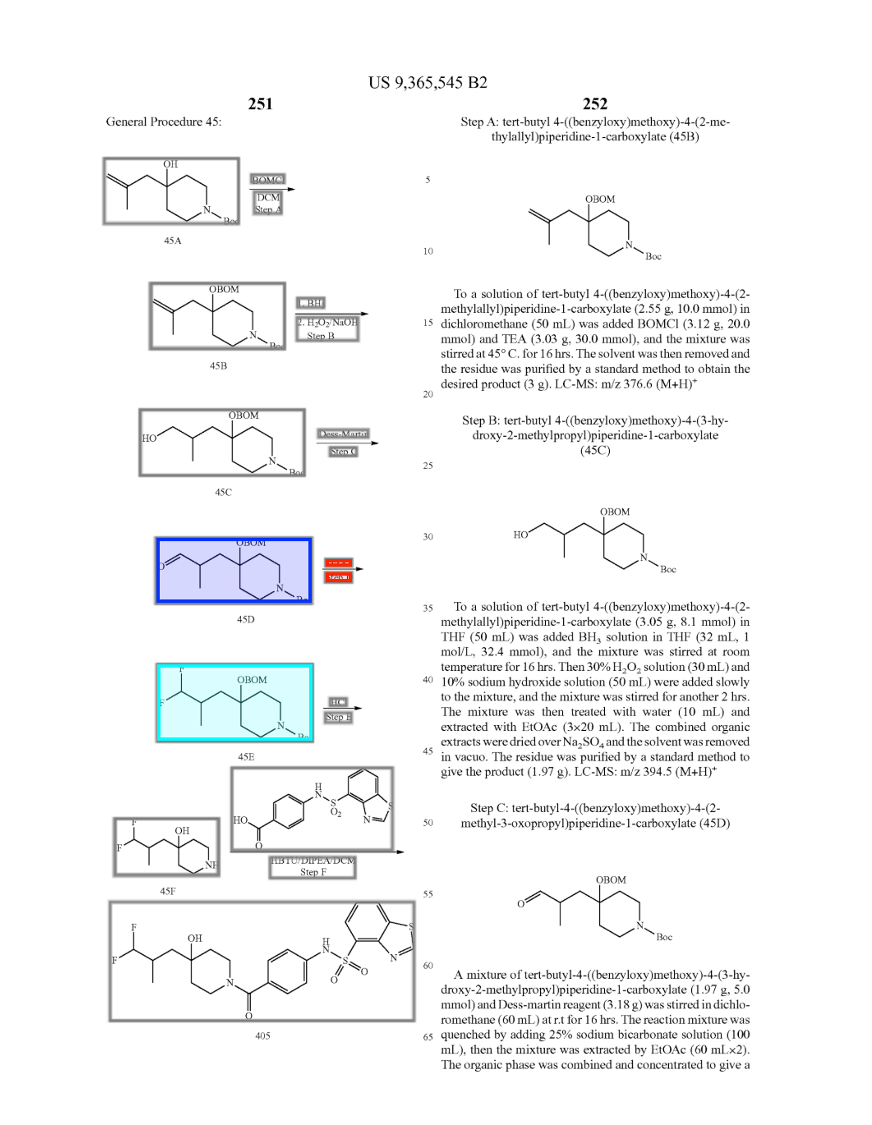}}
  \caption{Sample ViReact results from our testset. Dark blue, red and light blue indicate reactants, conditions and products, respectively.}
  \label{fig:vireact_testset2}
\end{figure}

In this section, we present qualitative results of MolMole on our testsets. 
\autoref{fig:videtect_testset} shows sample test pages with ViDetect inference results. The testset includes simple cases where the page contains a few molecules with clear boundaries, as well as more challenging cases where many molecules are present or densely packed within a table. 
\autoref{fig:vireact_testset}--\autoref{fig:vireact_testset2} show additional testsets with ViReact inference results. As shown, the testsets feature documents with complex layouts (e.g., two-column formats) and a variety of reaction diagrams, including multi-line and tree diagrams.
In all cases, MolMole accurately predicted molecules and reactions. Complete extraction results are available on the \href{https://lgai-ddu.github.io/molmole/}{MolMole project page}.

\subsubsection{MolMole Workflow}
\begin{figure}[h]
\centering
\fbox{\includegraphics[width=0.9\textwidth]{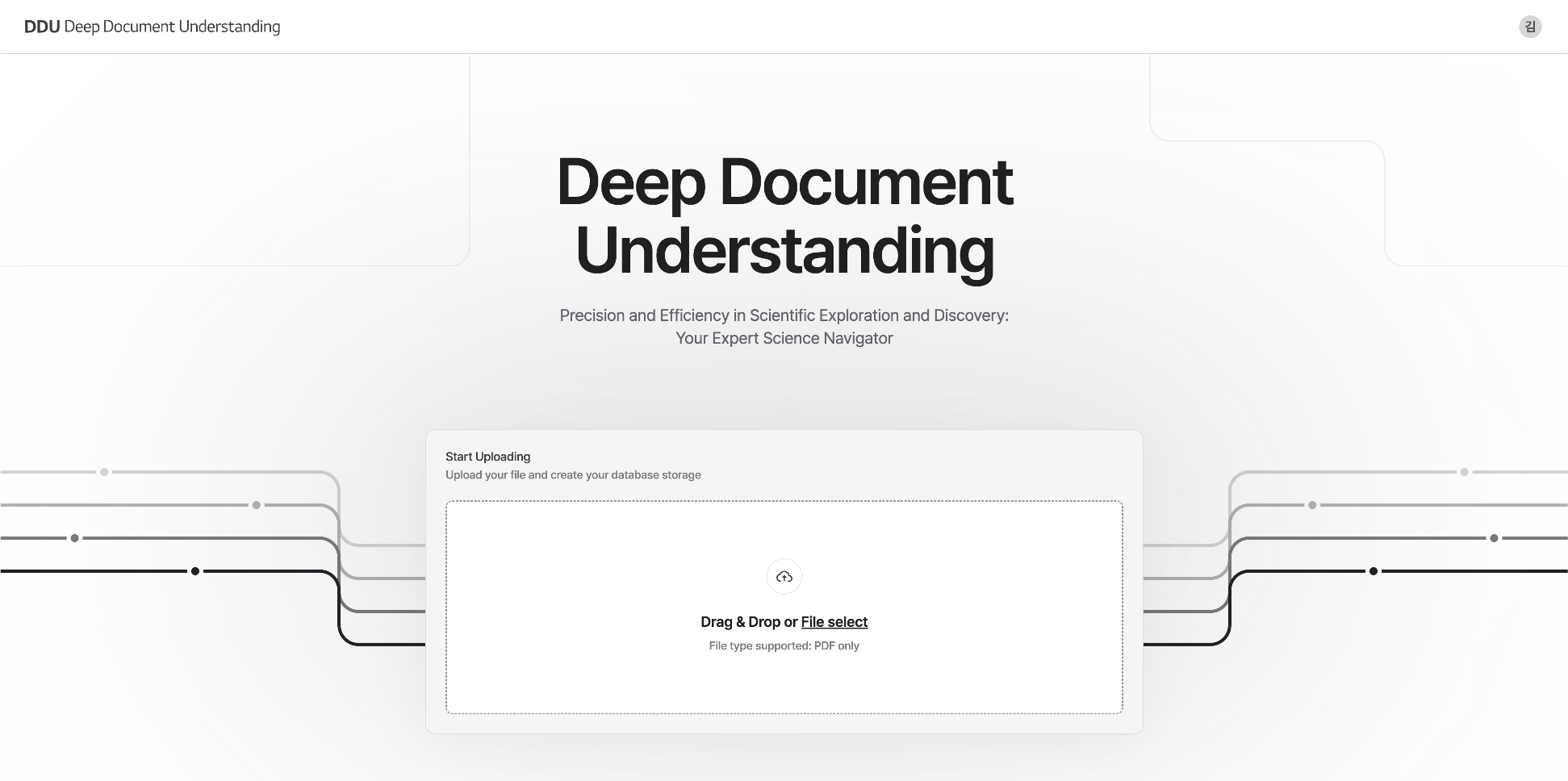}}
\caption{Upload your PDF file.}
\label{fig:workflow_first_page}
\end{figure}

\begin{figure}[h]
\centering
\fbox{\includegraphics[width=0.9\textwidth]{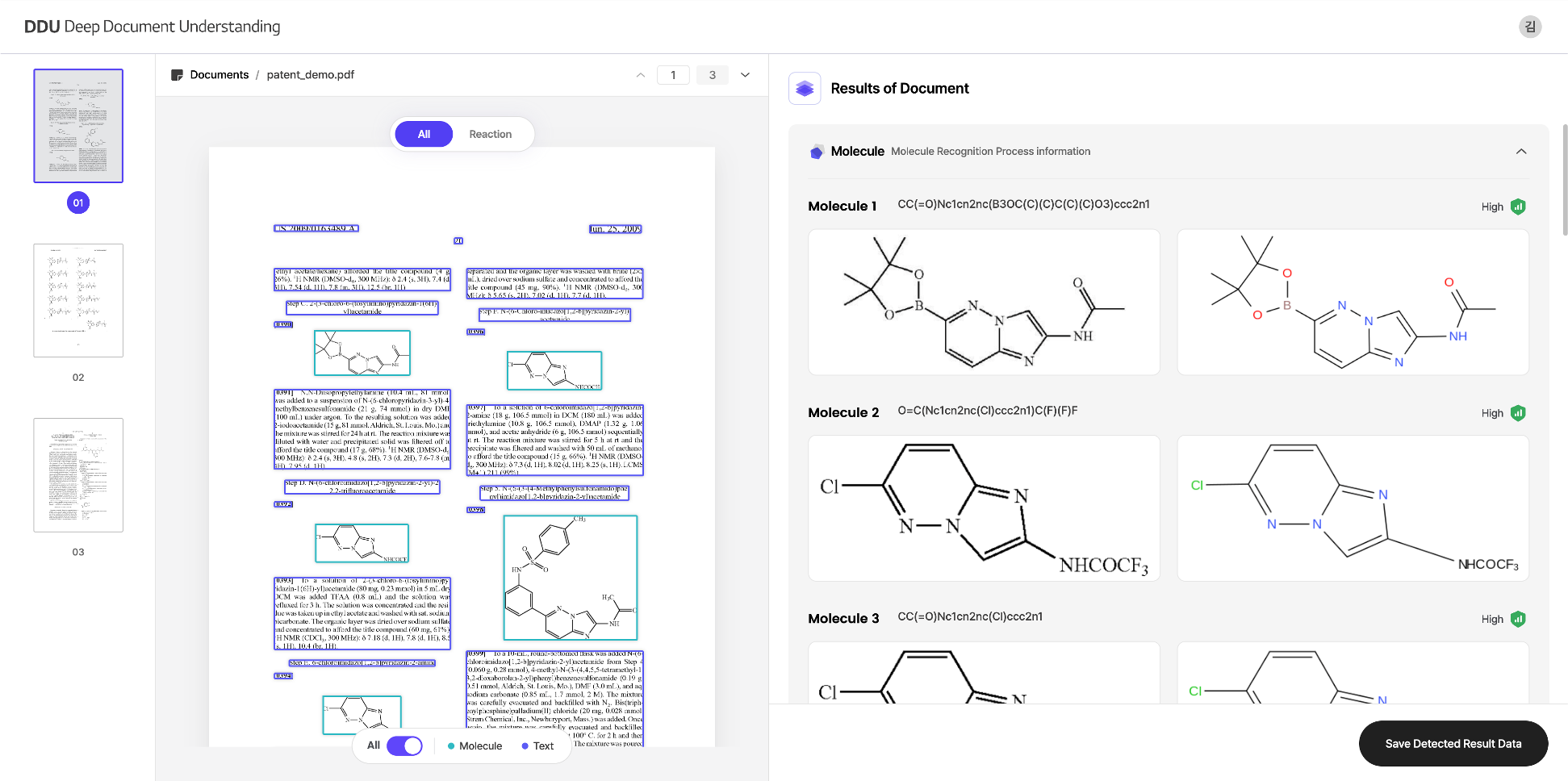}}
\caption{Processed results. On the left, you can check ViDetect detection results of molecular regions. On the right, you can check the molecule conversion results in SMILES through ViMore with confidence score among low, medium, high.}
\label{fig:workflow_results}
\end{figure}

\begin{figure}[h]
\centering
\fbox{\includegraphics[width=0.9\textwidth]{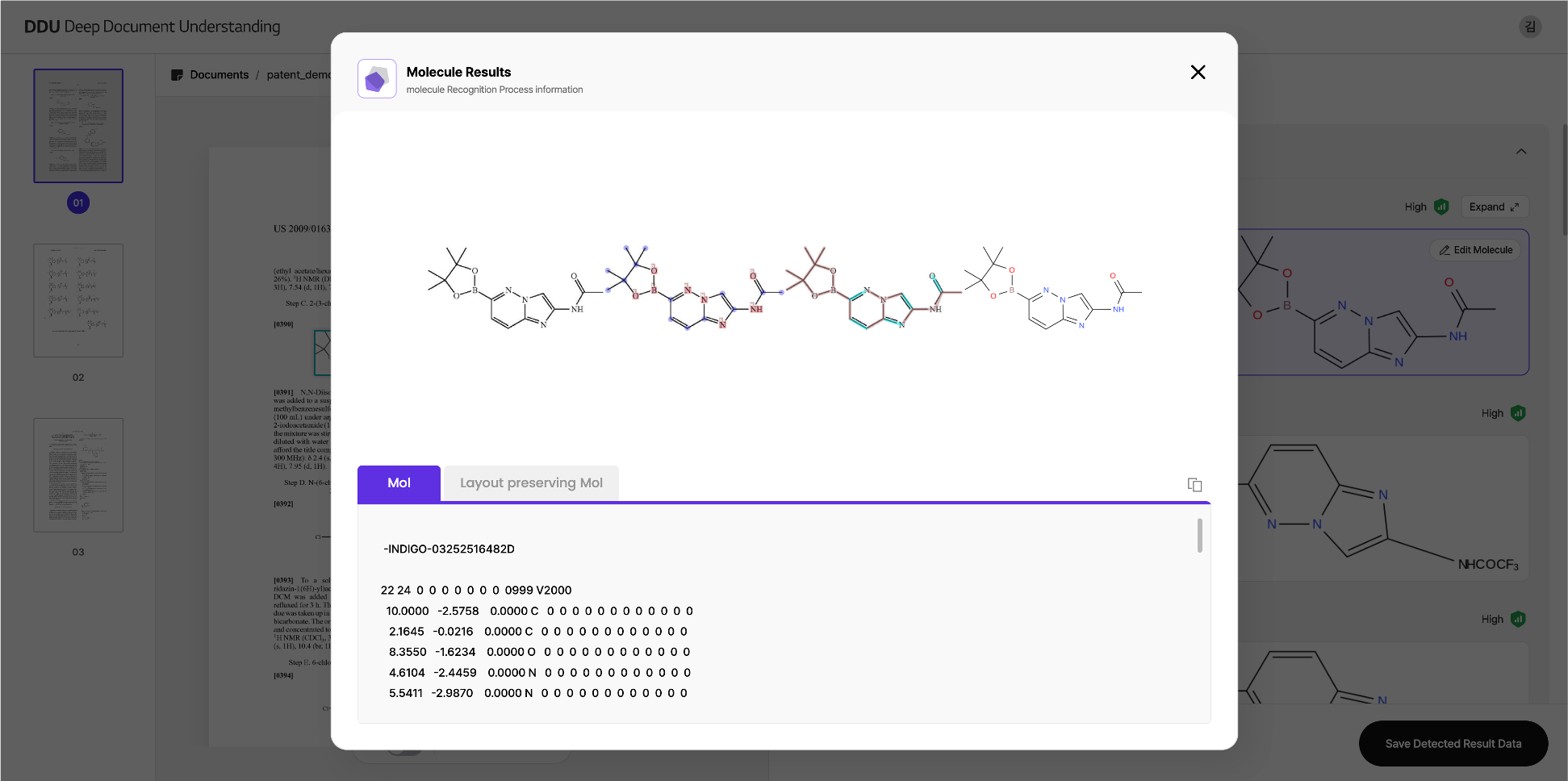}}
\caption{By clicking on expand, you can check the detailed model prediction results with MOLfile.}
\label{fig:workflow_molexpand}
\end{figure}

\begin{figure}[h]
\centering
\fbox{\includegraphics[width=0.9\textwidth]{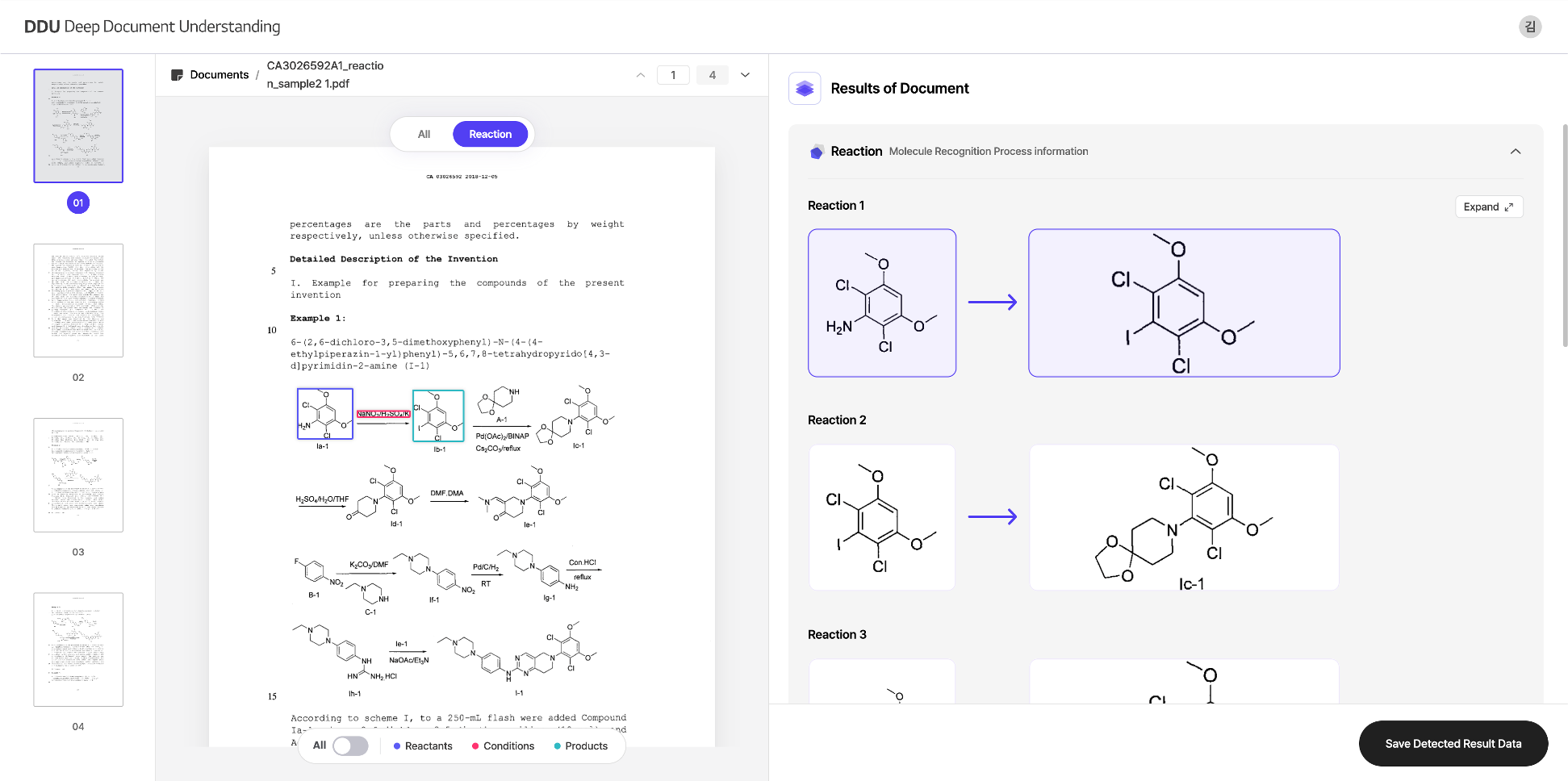}}
\caption{When there are reactions in the page, you can switch to Reaction mode. The detected reaction is visualized sequentially. The image is showing the first reaction detected in the page.}
\label{fig:workflow_reaction}
\end{figure}

\begin{figure}[h]
\centering
\fbox{\includegraphics[width=0.9\textwidth]{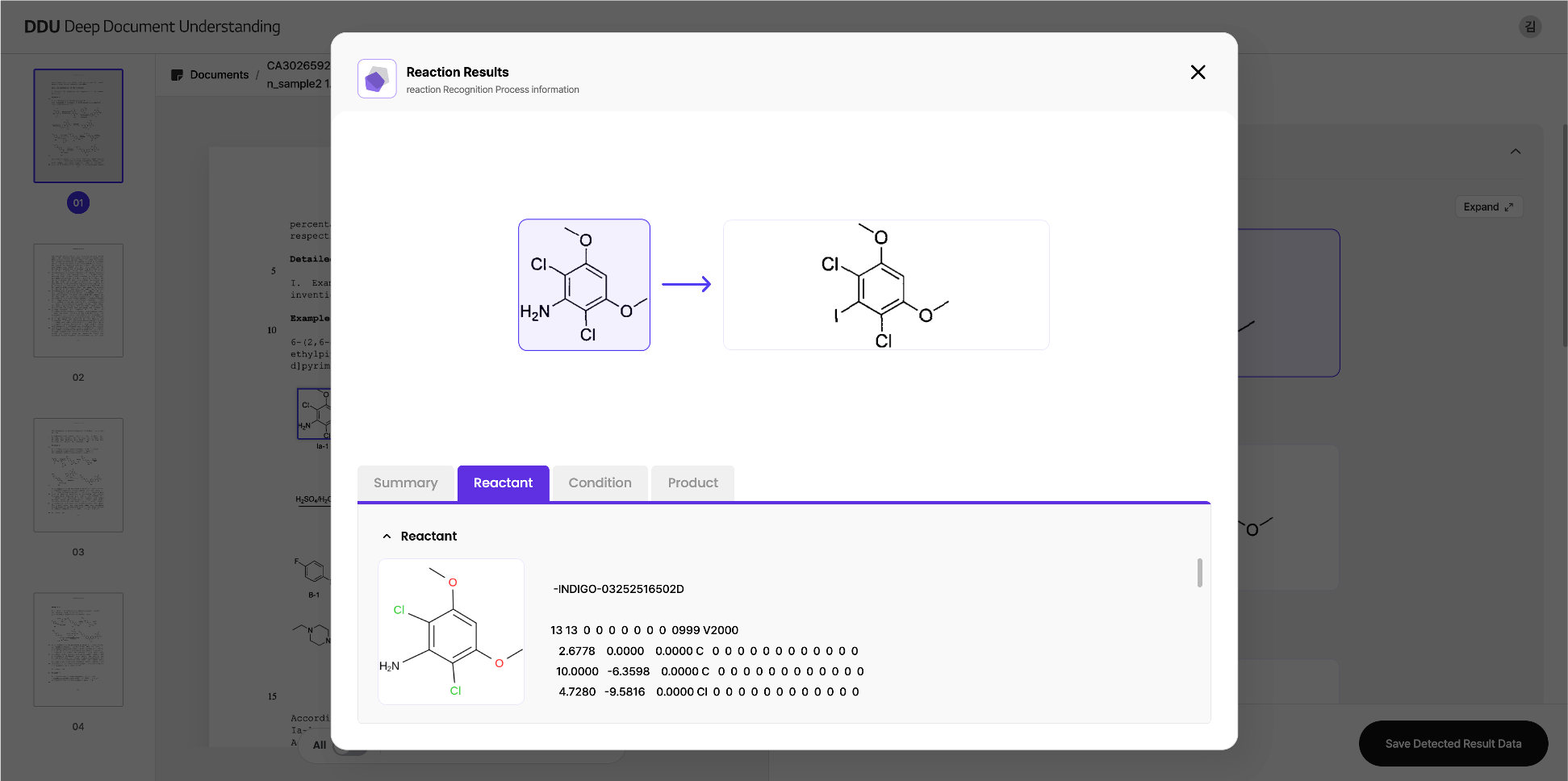}}
\caption{By clicking on expand, you can check the detected reactant, condition and product converted in MOLfile format. }
\label{fig:workflow_reaction_expand}
\end{figure}

To provide a clear understanding of how MolMole operates, Figures~\ref{fig:workflow_first_page}--\ref{fig:workflow_reaction_expand} present a step-by-step visualization of the extraction process. The screenshots illustrate the complete workflow—from document input, molecule detection, and reaction diagram parsing to structure conversion. A demo video is also available on the \href{https://lgai-ddu.github.io/molmole/}{MolMole project page}.
\setlength{\fboxsep}{0pt}     
\setlength{\fboxrule}{0.5pt}  

\clearpage
\bibliographystyle{abbrvnat}
\bibliography{egbib}

\begin{thebibliography}{15}
\providecommand{\natexlab}[1]{#1}
\providecommand{\url}[1]{\texttt{#1}}
\expandafter\ifx\csname urlstyle\endcsname\relax
  \providecommand{\doi}[1]{doi: #1}\else
  \providecommand{\doi}{doi: \begingroup \urlstyle{rm}\Url}\fi

\bibitem[Carion et~al.(2020)Carion, Massa, Synnaeve, Usunier, Kirillov, and Zagoruyko]{carion2020end}
N.~Carion, F.~Massa, G.~Synnaeve, N.~Usunier, A.~Kirillov, and S.~Zagoruyko.
\newblock End-to-end object detection with transformers.
\newblock In \emph{European conference on computer vision}, pages 213--229. Springer, 2020.

\bibitem[Dalby et~al.(1992)Dalby, Nourse, Hounshell, Gushurst, Grier, Leland, and Laufer]{dalby1992description}
A.~Dalby, J.~G. Nourse, W.~D. Hounshell, A.~K. Gushurst, D.~L. Grier, B.~A. Leland, and J.~Laufer.
\newblock Description of several chemical structure file formats used by computer programs developed at molecular design limited.
\newblock \emph{Journal of chemical information and computer sciences}, 32\penalty0 (3):\penalty0 244--255, 1992.

\bibitem[Fan et~al.(2024)Fan, Qian, Wang, Wang, Coley, and Barzilay]{fan2024openchemie}
V.~Fan, Y.~Qian, A.~Wang, A.~Wang, C.~W. Coley, and R.~Barzilay.
\newblock Openchemie: An information extraction toolkit for chemistry literature.
\newblock \emph{Journal of Chemical Information and Modeling}, 64\penalty0 (14):\penalty0 5521--5534, 2024.

\bibitem[Heller et~al.(2015)Heller, McNaught, Pletnev, Stein, and Tchekhovskoi]{heller2015inchi}
S.~R. Heller, A.~McNaught, I.~Pletnev, S.~Stein, and D.~Tchekhovskoi.
\newblock Inchi, the iupac international chemical identifier.
\newblock \emph{Journal of cheminformatics}, 7:\penalty0 1--34, 2015.

\bibitem[Lin et~al.(2014)Lin, Maire, Belongie, Hays, Perona, Ramanan, Doll{\'a}r, and Zitnick]{lin2014microsoft}
T.-Y. Lin, M.~Maire, S.~Belongie, J.~Hays, P.~Perona, D.~Ramanan, P.~Doll{\'a}r, and C.~L. Zitnick.
\newblock Microsoft coco: Common objects in context.
\newblock In \emph{Computer vision--ECCV 2014: 13th European conference, zurich, Switzerland, September 6-12, 2014, proceedings, part v 13}, pages 740--755. Springer, 2014.

\bibitem[Morin et~al.(2023)Morin, Danelljan, Agea, Nassar, Weber, Meijer, Staar, and Yu]{morin2023molgrapher}
L.~Morin, M.~Danelljan, M.~I. Agea, A.~Nassar, V.~Weber, I.~Meijer, P.~Staar, and F.~Yu.
\newblock Molgrapher: graph-based visual recognition of chemical structures.
\newblock In \emph{Proceedings of the IEEE/CVF International Conference on Computer Vision}, pages 19552--19561, 2023.

\bibitem[Qian et~al.(2023{\natexlab{a}})Qian, Guo, Tu, Coley, and Barzilay]{qian2023rxnscribe}
Y.~Qian, J.~Guo, Z.~Tu, C.~W. Coley, and R.~Barzilay.
\newblock Rxnscribe: a sequence generation model for reaction diagram parsing.
\newblock \emph{Journal of chemical information and modeling}, 63\penalty0 (13):\penalty0 4030--4041, 2023{\natexlab{a}}.

\bibitem[Qian et~al.(2023{\natexlab{b}})Qian, Guo, Tu, Li, Coley, and Barzilay]{qian2023molscribe}
Y.~Qian, J.~Guo, Z.~Tu, Z.~Li, C.~W. Coley, and R.~Barzilay.
\newblock Molscribe: robust molecular structure recognition with image-to-graph generation.
\newblock \emph{Journal of Chemical Information and Modeling}, 63\penalty0 (7):\penalty0 1925--1934, 2023{\natexlab{b}}.

\bibitem[Rajan et~al.(2020)Rajan, Brinkhaus, Zielesny, and Steinbeck]{rajan2020review}
K.~Rajan, H.~O. Brinkhaus, A.~Zielesny, and C.~Steinbeck.
\newblock A review of optical chemical structure recognition tools.
\newblock \emph{Journal of Cheminformatics}, 12:\penalty0 1--13, 2020.

\bibitem[Rajan et~al.(2021)Rajan, Brinkhaus, Sorokina, Zielesny, and Steinbeck]{rajan2021decimer}
K.~Rajan, H.~O. Brinkhaus, M.~Sorokina, A.~Zielesny, and C.~Steinbeck.
\newblock Decimer-segmentation: automated extraction of chemical structure depictions from scientific literature.
\newblock \emph{Journal of cheminformatics}, 13:\penalty0 1--9, 2021.

\bibitem[Rajan et~al.(2023)Rajan, Brinkhaus, Agea, Zielesny, and Steinbeck]{rajan2023decimer}
K.~Rajan, H.~O. Brinkhaus, M.~I. Agea, A.~Zielesny, and C.~Steinbeck.
\newblock Decimer. ai: an open platform for automated optical chemical structure identification, segmentation and recognition in scientific publications.
\newblock \emph{Nature communications}, 14\penalty0 (1):\penalty0 5045, 2023.

\bibitem[Shen et~al.(2021)Shen, Zhang, Dell, Lee, Carlson, and Li]{shen2021layoutparser}
Z.~Shen, R.~Zhang, M.~Dell, B.~C.~G. Lee, J.~Carlson, and W.~Li.
\newblock Layoutparser: A unified toolkit for deep learning based document image analysis.
\newblock In \emph{Document Analysis and Recognition--ICDAR 2021: 16th International Conference, Lausanne, Switzerland, September 5--10, 2021, Proceedings, Part I 16}, pages 131--146. Springer, 2021.

\bibitem[Weininger(1988)]{weininger1988smiles}
D.~Weininger.
\newblock Smiles, a chemical language and information system. 1. introduction to methodology and encoding rules.
\newblock \emph{Journal of chemical information and computer sciences}, 28\penalty0 (1):\penalty0 31--36, 1988.

\bibitem[Wilary and Cole(2023)]{wilary2023reactiondataextractor}
D.~M. Wilary and J.~M. Cole.
\newblock Reactiondataextractor 2.0: a deep learning approach for data extraction from chemical reaction schemes.
\newblock \emph{Journal of Chemical Information and Modeling}, 63\penalty0 (19):\penalty0 6053--6067, 2023.

\bibitem[Zhang et~al.(2022)Zhang, Li, Liu, Zhang, Su, Zhu, Ni, and Shum]{zhang2022dino}
H.~Zhang, F.~Li, S.~Liu, L.~Zhang, H.~Su, J.~Zhu, L.~M. Ni, and H.-Y. Shum.
\newblock Dino: Detr with improved denoising anchor boxes for end-to-end object detection.
\newblock \emph{arXiv preprint arXiv:2203.03605}, 2022.

\end{thebibliography}

\end{document}